\documentclass[11pt]{article}

\usepackage[final]{acl}

\usepackage{iftex}
\usepackage{latexsym}

\ifPDFTeX
  \usepackage{times}
  \usepackage[T1]{fontenc}
  \usepackage[utf8]{inputenc}
  \usepackage{inconsolata}
  \newcommand{\rubriccodefont}{\ttfamily}
\else
  \usepackage{fontspec}
  \newfontfamily\rubriccodefont{DejaVuSansMono}[
    Extension=.ttf,
    UprightFont=*,
    BoldFont=*-Bold,
    ItalicFont=*-Oblique,
    BoldItalicFont=*-BoldOblique
  ]
\fi

\usepackage{microtype}
\usepackage{graphicx}
\usepackage{amsmath}
\usepackage{amssymb}
\usepackage{booktabs}
\usepackage{multirow}
\usepackage{xcolor}
\usepackage[most]{tcolorbox}
\usepackage{colortbl}
\usepackage{adjustbox}
\usepackage{makecell}
\usepackage{tabularx}
\usepackage{stfloats}

\title{RubricRM: Generative Reward Modeling via Dynamic Rubrics for Image Generation and Editing}

\author{
  \textbf{Zijian Kan\textsuperscript{1,2},}
  \textbf{Wei Wang\textsuperscript{2},}
  \textbf{Long Luo\textsuperscript{2},}
  \textbf{Bing Zhao\textsuperscript{3,}}\thanks{Corresponding author.},
  \textbf{Xuan Ren\textsuperscript{2},}\\
  \textbf{Weixu Qiao\textsuperscript{2},}
  \textbf{Wenbo Li\textsuperscript{2},}
  \textbf{Hu Wei\textsuperscript{2},}
  \textbf{Lin Qu\textsuperscript{2}}\\
  \normalfont
  \textsuperscript{1}Peking University \quad
  \textsuperscript{2}Alibaba Group \quad
  \textsuperscript{3}Alibaba DAMO Academy \\
  \texttt{kanzijian26@stu.pku.edu.cn} \quad
  \texttt{xiongdao@alibaba-inc.com}
}

\begin{document}
\maketitle

\begin{abstract}
Reward models play an essential role in aligning visual generative models, yet most existing visual reward models use a single scalar score or rely on fixed criteria that cannot adapt to different instructions. This limits both interpretability and task sensitivity, especially for text-to-image generation and instruction-based image editing, where different inputs require different evaluation dimensions. We propose \mbox{RubricRM}, a pairwise generative reward modeling framework that first produces an input-specific rubric with evaluation dimensions, weights, and scoring criteria, and then applies the rubric to score candidate images. We train dedicated RubricRM models for text-to-image generation and image editing using a two-stage training pipeline: supervised fine-tuning teaches the model the rubric-based scoring paradigm, while GRPO further improves scoring through fine-grained dimension-level rewards. Experiments on multiple generation and editing benchmarks show that RubricRM outperforms existing specialized reward models and remains competitive with strong proprietary MLLM judges despite using smaller backbones. Our models, data, and code are available at \url{https://github.com/zijiankan/RubricRM}.
\end{abstract}

\section{Introduction}

\begin{figure*}[t]
\centering
\includegraphics[width=\textwidth]{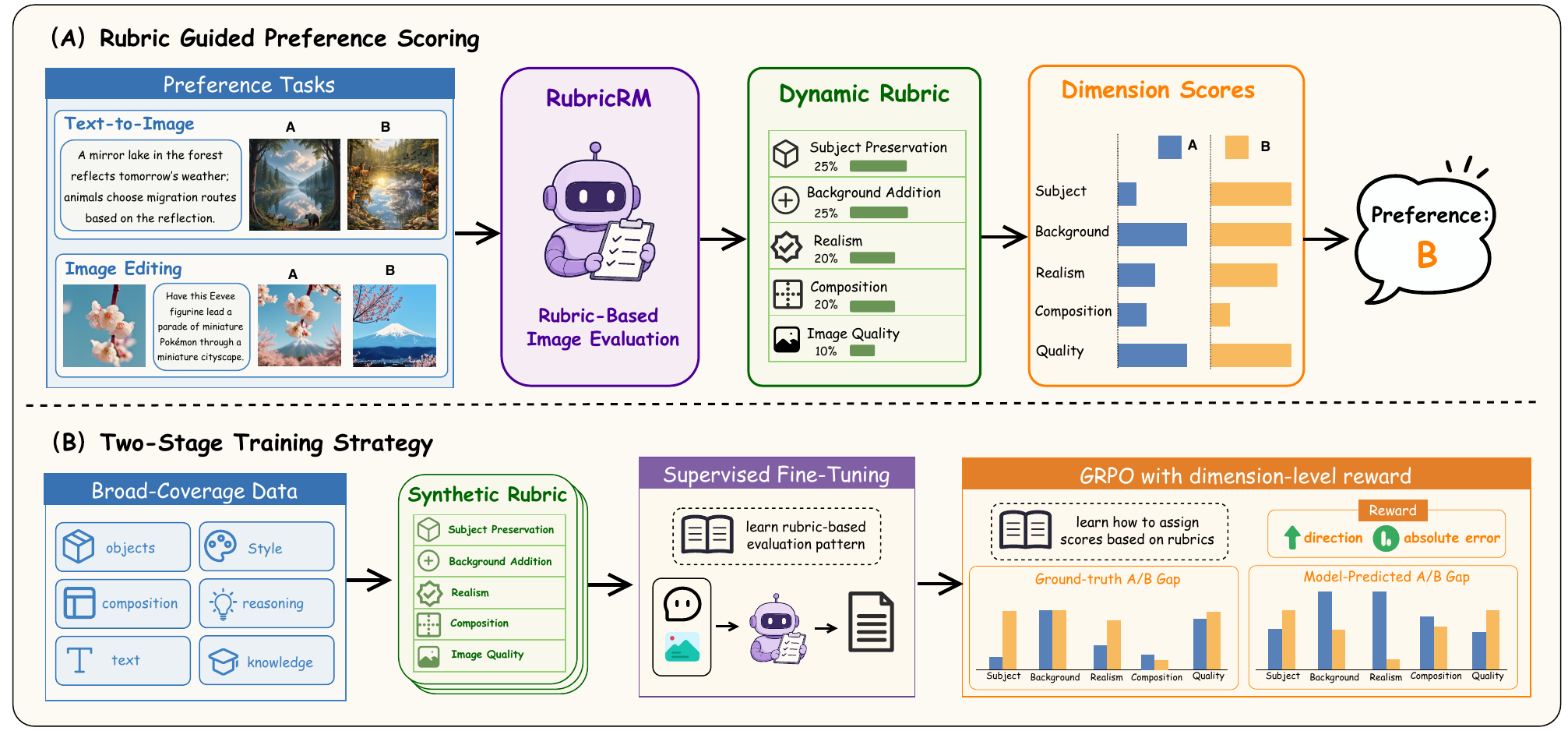}
\caption{\textbf{Inference and training pipeline of RubricRM.} 
(A) RubricRM supports both text-to-image and image-editing preference selection. At inference time, the model first dynamically generates a rubric conditioned on the input, including scoring criteria and corresponding weights, and then scores each image along each dimension based on the rubric, all within a single inference pass.
(B) Training is conducted in two stages. The model is first supervised fine-tuned on synthesized rubric data to learn the paradigm of rubric generation followed by scoring, and is then further optimized with GRPO using dimension-level rewards.\vspace{-0.6cm}}

\label{fig:overview}
\end{figure*}

Reward models are essential for the post-training alignment of visual generative models, as they convert human preferences into learnable reward signals \cite{ouyang2022training,black2023training,fan2023dpok,wallace2024diffusion}.

Existing reward models fall into several categories. CLIP-based metrics \cite{hessel2021clipscore} offer efficient image-text alignment signals yet exhibit limited sensitivity to fine-grained compositional semantics. Models such as ImageReward \cite{xu2023imagereward}, HPS v2 \cite{wu2023human,wu2023hpsv2}, HPS v3 \cite{ma2025hpsv3}, PickScore \cite{kirstain2023pickapic}, and EditReward \cite{wu2025editreward} are trained on large-scale preference data but produce scalar scores that offer little interpretability. Recent generative reward models employ multimodal large language models to generate evaluation rationales \cite{ku2024viescore,wang2025think}. Yet these approaches typically adopt predefined and fixed evaluation dimensions, whereas different instructions naturally demand distinct assessment criteria. For example, realistic prompts prioritize physical plausibility, whereas creative prompts value novelty, highlighting the need for adaptive evaluation rubrics.

To address this limitation, we introduce automated rubric generation for evaluating text-to-image generation and image editing. Our central idea is to use a rubric as an intermediate evaluation protocol between the instruction and the preference decision. The rubric specifies what aspects should be evaluated, how much each aspect matters, and how each score level should be interpreted, allowing the evaluation criteria to adapt to the input while keeping the scoring process inspectable.

Based on this concept, we propose Rubric-based Reward Model (RubricRM), a generative reward modeling framework for text-to-image generation and image editing. Given an instruction and two candidate outputs, RubricRM constructs an input-specific rubric covering task intent, atomic evaluation dimensions, dimension weights, and scoring criteria. It then scores both candidates along each dimension and aggregates the weighted dimension scores to derive the final preference.

RubricRM is trained via a two-stage pipeline that targets two capabilities: learning the rubric-based scoring pattern and refining scoring accuracy. In the supervised fine-tuning stage, the model imitates teacher-synthesized rubric trajectories to learn how to decompose an instruction into intent, dimensions, weights, and graded criteria. In the subsequent reinforcement learning stage, dimension-level rewards align the model's predicted score differences with reference score differences on each rubric dimension and apply stronger penalties when the preference direction is reversed.

Learning dynamic rubrics requires broad data coverage, since a narrow training distribution can bias the learned rubrics and impair generalization. To this end, we collect multiple open-source preference datasets and analyze instruction types in text-to-image generation and image editing. We design prompt and editing-instruction classification criteria to support taxonomy annotation, balanced sampling, and the identification of underrepresented categories. For these categories, we synthesize supplementary prompts, generate candidate images, and collect expert preference annotations. We then synthesize rubric trajectories for each preference sample, yielding 31.8K text-to-image and 30.7K image-editing training samples.

We train RubricRM on Qwen3.5-4B and Qwen3.5-9B backbone models \cite{qwen35blog}. For text-to-image generation, RubricRM-Gen-9B attains accuracies of 72.00, 74.12, and 84.45 on MMRB2, GenAI-Bench, and GenAI-Bench-Verified, respectively. For image editing, RubricRM-Edit-9B reaches 75.40, 46.40, and 85.64 on MMRB2, EditReward-ERB, and EditScore-ERB, outperforming existing reward-model baselines across both tasks.

In summary, our key contributions are as follows:
\begin{itemize}
    \item We introduce a dynamic rubric-based reward modeling paradigm for text-to-image generation and image editing and develop specialized models for both tasks.
    \item We construct rubric trajectories from preference data and train RubricRM with a two-stage procedure that combines rubric-trajectory SFT with dimension-level GRPO.
    \item We obtain strong results on multiple reward benchmarks, outperforming specialized reward-model baselines across generation and editing tasks.
\end{itemize}

\section{Related Work}

\paragraph{Text-to-image generation evaluation.}
Text-to-image evaluation has moved from global alignment metrics to preference-aligned reward modeling. CLIPScore \cite{hessel2021clipscore} measures image-text similarity with CLIP \cite{radford2021learning}, but it is unreliable for compositional semantics \cite{huang2023t2icompbench,jiang2024genai}. Preference-trained models such as ImageReward \cite{xu2023imagereward}, HPS v2 \cite{wu2023human,wu2023hpsv2}, HPS v3 \cite{ma2025hpsv3}, and PickScore \cite{kirstain2023pickapic} improve human alignment, but their standard interface is still a scalar reward, which provides limited evidence for the decision. VisionReward \cite{xu2024visionreward} introduces multi-dimensional scoring, but the dimensions and weights are predefined rather than generated from the instruction.

Recent generative reward models use MLLMs to produce evaluative reasoning: UnifiedReward \cite{wang2025unified} supports multiple multimodal reward tasks, and UnifiedReward-Think \cite{wang2025think} adds explicit reasoning while relying on fixed scoring criteria. UnifiedReward-Flex \cite{wang2026flex} further introduces personalized, context-adaptive assessment for vision generation, but it does not explicitly specify the detailed requirements of each scoring criterion or assign weights to the criteria.

\paragraph{Image editing evaluation.}
Instruction-guided image editing \cite{brooks2023instructpix2pix} sharpens the need for adaptive criteria: an edit must satisfy the instruction without corrupting irrelevant content. EditReward \cite{wu2025editreward} learns from over 200K editing preference pairs, but remains a discriminative scalar reward model. EditScore \cite{luo2025editscore} builds high-fidelity generative reward models and uses them for online RL in image editing, but it does not define instruction-specific rubrics with explicit weights and scoring levels. RubricRM targets this layer between free-form rationale and scalar preference, where the model must decide which aspects of edit execution, source preservation, and visual quality should be emphasized for each instruction.

\paragraph{Rubric-based evaluation.}
Rubrics have recently emerged as a mechanism for making evaluation criteria explicit. AutoRubric \cite{autorubric2026} generates task-adapted rubrics, RubricHub \cite{rubrichub2026} constructs large-scale rubric resources, and OpenRubrics \cite{openrubrics2025} studies scalable synthetic rubric generation for reward modeling and alignment. In multimodal settings, AutoRubric-R1V \cite{jia2025autorubric} uses rubric-based process rewards for multimodal reasoning RL, while Proxy-GRM \cite{qiu2026rationale} learns transferable rubrics for vision-language response preference judgment through proxy-guided critique. AutoRubric-T2I \cite{kao2026autorubrict2i} learns a compact, globally shared set of weighted natural-language rules from preference data; an external frozen VLM scores each image against these rules, whose scores are aggregated for text-to-image evaluation and downstream generation-model RL. Auto-Rubric as Reward \cite{tian2026autorubricreward} instead uses frozen VLMs to generate and verify prompt-conditioned binary criteria, then makes a holistic rubric-conditioned preference decision for text-to-image generation and image editing. Unlike these approaches, RubricRM trains a standalone, self-contained generative reward model that produces input-specific dimensions, normalized weights, graded scoring criteria, and explicit per-dimension scores for both candidates in a single end-to-end inference pass.

\section{RubricRM}
\label{sec:method}
This section first defines the rubric-based visual reward modeling task, then describes the rubric trajectory data, and finally presents the two-stage training procedure.

\subsection{Task Definition}

We formally define the task of rubric-based visual reward modeling as follows. For text-to-image generation, the input consists of a text prompt $q$ and two candidate images $I_A, I_B$. For image editing, $q$ additionally includes a source image $I_{\text{src}}$ and an editing instruction. The goal is to predict a preference label $y \in \{A, B\}$ indicating which image better satisfies the given instruction.

RubricRM produces a structured rubric together with dimension-level assessments for both candidates in a single forward pass. Formally, given $q$ and the two candidate images, the model outputs a rubric
\begin{equation}
\mathcal{R} = \{ (d_i, w_i, \mathcal{L}_i) \}_{i=1}^{N}
\end{equation}
and, for each image $I_k$ with $k \in \{A, B\}$, a set of dimension scores $\{s_i^k\}_{i=1}^N$. Here $N$ is the number of atomic evaluation dimensions, $d_i$ is the description of the $i$-th dimension, $w_i \in [0,1]$ is its relative weight with $\sum_{i=1}^N w_i = 1$, and $\mathcal{L}_i$ gives the specific scoring criteria for this dimension. Each $s_i^k \in [0,4]$ is determined by referencing $d_i$ and $\mathcal{L}_i$. The rubric thus serves as an interpretable blueprint that specifies \textit{what to evaluate}, \textit{how important each aspect is}, and \textit{what each score level means}.

The overall score for image $I_k$ is then computed as the weighted sum
\begin{equation}
S(I_k) = \sum_{i=1}^{N} w_i \cdot s_i^k
\end{equation}

The final preference decision is made by comparing $S(I_A)$ and $S(I_B)$, with the higher-scoring image being selected as the preferred one. Complete running cases for text-to-image generation and image editing are provided in Appendices~\ref{sec:appendix_running_t2i} and~\ref{sec:appendix_running_edit}.

\subsection{Rubric Trajectory Data}

\paragraph{Classification-Driven Balanced Sampling.} RubricRM relies on dynamically generated scoring criteria, so the training data should cover diverse instruction types and visual scenarios. To improve coverage, we design a classification-driven balanced sampling strategy.

Specifically, we analyze the subtasks in text-to-image generation and image editing, and develop a set of prompt classification criteria from multiple dimensions. We then employ DeepSeek-V3.2 for automatic category annotation. Since the same image may belong to different categories when viewed from different dimensions (e.g., a portrait photo falls under ``portrait'' from the object perspective, but may be classified as ``close-up'' from the photographic style perspective), the model is allowed to return multiple labels for each text prompt or editing instruction. This classification strategy serves three purposes: guiding balanced sampling from open-source data, identifying underrepresented categories for targeted supplementary synthesis, and facilitating the split of SFT and RL data.

During data splitting, most text-to-image samples carry multiple labels and cannot be assigned to a single category. We therefore represent each sample as a set of labels and adopt multi-label stratified sampling to balance coverage across second-level labels. Table~\ref{tab:data_stats} summarizes the training data. Appendix~\ref{sec:appendix_data} gives additional details on the RL split construction.

\paragraph{Text-to-Image Preference Data.} The text-to-image data comprise open-source datasets and our additionally annotated data. Each sample consists of a generation prompt, two candidate images, and a preference label. The open-source data are sourced from HPD v3 \cite{ma2025hpsv3}, OIP\footnote{\url{https://huggingface.co/datasets/data-is-better-together/open-image-preferences-v1-binarized}}, and EvalMuse-40K \cite{han2026evalmuse40k}. We observed that some categories are underrepresented in the open-source data. Following the prompt synthesis procedure of \citet{wang2026everything}, we use DeepSeek-V3.2 to synthesize supplementary prompts for these categories and manually screen the generated prompts for quality. We then render the retained prompts with ten image generation models, including GPT-Image, Gemini-3.1-Pro, Seedream-5.0, and Wan-2.6. Three human experts rate each image on a 10-point scale, and we retain only pairs whose score difference exceeds 5 points with unanimous preference agreement. The final text-to-image training set contains 31,835 samples.

\paragraph{Image Editing Preference Data.} The image editing data are built upon EditReward-Data \cite{wu2025editreward}. Each sample contains a source image, an editing instruction, two candidate edited results, and a preference label. Since EditReward-Data already provides relatively balanced instruction coverage, we do not synthesize additional editing prompts. The final image editing training set comprises 30,695 samples.

\paragraph{Distribution Analysis.}
Classification results show that in text-to-image generation, portraits and realistic photographic styles account for a relatively high proportion, whereas categories such as text rendering (6.45\%), logical reasoning (8.98\%), and world knowledge (15.83\%) are less prevalent. Our balanced sampling strategy is designed to retain coverage of these long-tail categories. In image editing, object and attribute editing are the most common, while reasoning-based editing (8.70\%), text editing (11.17\%), and style editing (10.18\%) appear less frequently. Figure~\ref{fig:appendix_primary_coverage} visualizes primary-category coverage for both tasks, and Tables~\ref{tab:t2i_secondary_labels}--\ref{tab:edit_secondary_labels} list the complete second-level taxonomy coverage.
\begin{table}[t]
\centering
\caption{Training data statistics for rubric learning. P/S denotes primary/secondary labels. }
\label{tab:data_stats}
\footnotesize
\setlength{\tabcolsep}{5pt}
\renewcommand{\arraystretch}{1.08}
\begin{tabular}{@{}lcc@{}}
\toprule
\textbf{Metric} & \textbf{Text-to-Image} & \textbf{Image Editing} \\
\midrule
\#Pairs & 31,835 & 30,695 \\
SFT/RL & 16,835/15,000 & 15,695/15,000 \\
Taxonomy (P/S) & 6/42 & 12/30 \\
Avg. tags & 3.3 & 1.4 \\
Multi-tag (\%) & 94.9 & 31.8 \\
Avg. dim. & 3.9 & 3.2 \\
A/B (\%) & 51.6/48.4 & 55.7/44.3 \\
\bottomrule
\end{tabular}
\end{table}

\paragraph{Rubric Trajectory Synthesis.} Training RubricRM requires jointly learning rubric generation and dimension-level scoring, yet preference labels alone are insufficient to provide these supervision signals. Since manually authoring full rubrics with scoring for large-scale data does not scale, we employ \mbox{Gemini 3.1 Pro} as the teacher model to synthesize training trajectories.

We use label-conditioned trajectory synthesis. To prevent synthesized trajectory quality from being bottlenecked by the teacher model's standalone preference judgment, we provide the human preference label $y_j$ to the teacher together with the prompt and candidate images. Thus, the teacher model is not used as an independent preference annotator; instead, it converts an existing human preference label into a structured rubric trajectory that supervises both criterion formulation and dimension-level scoring. The label anchors the trajectory to the human preference while still requiring the teacher model to specify evaluation dimensions, assign weights, define scoring criteria, and provide dimension-level evidence.

Concretely, given the text prompt $q_j$, the candidate images $I_A^j, I_B^j$, and the preference label $y_j$, the teacher model is prompted to generate four fields: task intent analysis, scoring dimensions with weight assignments, graded scoring descriptors for each dimension, and dimension-level assessments with scores for both candidate images.

\paragraph{Trajectory Quality Control.}
We apply a two-step quality control procedure after synthesis. First, we conduct structured parsing and discard trajectories with missing dimensions, out-of-range scores, weights that do not sum to 100\%, or those from which a final preference cannot be parsed. Second, we verify that the weighted total scores of the synthesized trajectory yield a preference direction consistent with the label $y_j$; if the teacher model's final preference $\hat{y}_j$ contradicts the label, the sample is removed.

To characterize the synthesized rubrics, Appendix~\ref{sec:appendix_data} further aggregates teacher-generated evaluation dimensions into semantic groups for text-to-image generation and image editing in Tables~\ref{tab:t2i_rubric_atoms} and~\ref{tab:edit_rubric_atoms}. Appendix~\ref{sec:appendix_rubric_analysis} examines rubric stability, diversity, and instruction-type adaptation, while Appendix~\ref{sec:appendix_teacher_sensitivity} reports a controlled analysis of sensitivity to the teacher family.

\subsection{Two-Stage Training}

We train RubricRM on the Qwen3.5-4B and Qwen3.5-9B backbone models for text-to-image generation and image editing, respectively. Both tasks follow the same training methodology: first, supervised fine-tuning enables the model to acquire the rubric-based scoring paradigm; then, reinforcement learning with fine-grained dimension-level rewards further refines the scoring accuracy.

\subsubsection{Stage 1: Supervised Fine-Tuning} Let $\mathcal{D}_{\text{SFT}} = \{(x_j, t_j)\}_{j=1}^{M_{\text{SFT}}}$ denote the filtered SFT dataset. Each input $x_j = (q_j, I_A^j, I_B^j)$ consists of a text prompt $q_j$ and two candidate images, and $t_j$ is the complete target trajectory synthesized by the teacher model, which sequentially contains task intent analysis, scoring dimensions with weights, graded scoring descriptors for each dimension, and dimension-level assessments with scores for both candidate images.

The model is optimized using the following loss function:
\begin{equation}
\mathcal{L}_{\text{SFT}} = -\sum_{j} \sum_{t=1}^{T_j} \log \pi_\theta(t_{j,t} \mid x_j, t_{j,<t})
\end{equation}

The core objective of this stage is for the model to learn the complete rubric-based judgment pipeline: decomposing an instruction into intent, selecting appropriate scoring dimensions and assigning weights, defining scoring criteria for each dimension, and producing dimension-level analysis leading to a final preference decision.

\subsubsection{Stage 2: GRPO with Dimension-Level Rewards}

\paragraph{Motivation.}
The SFT stage teaches the model a structured rubric-based evaluation pipeline, including generating evaluation dimensions, assigning weights, and producing dimension-level scores. However, trajectory imitation alone does not directly optimize final preference accuracy, and the model may still produce inaccurate dimension scores when it applies a rubric to fine-grained candidate comparison. We therefore introduce GRPO after SFT and use dimension-level rewards to optimize scoring behavior conditioned on a rubric.

Directly rewarding only the final preference provides a sparse signal and may assign the same reward to trajectories with substantially different scoring quality. We instead define rewards at a finer, dimension level, so that the model must not only select the correct preference but also align its score differences with the reference trajectory on each dimension.

\paragraph{Fixed-Rubric Rollout.}
Dimension-level rewards require the model-produced scoring dimensions to correspond one-to-one with the reference trajectory. During RL, we therefore fix the synthesized rubric for each sample and roll out only the subsequent dimension-level analysis and scores. Concretely, we split each teacher trajectory into a rubric part $\mathcal{R}_j$ and a scoring part $\mathcal{J}_j$. During training, $\mathcal{R}_j$ is treated as a given evaluation protocol, so rollouts for the same sample score under consistent dimensions, weights, and criteria. The predicted dimension scores can then be compared directly against the reference scores, allowing GRPO to focus on improving preference scoring accuracy conditioned on a rubric. This fixed-rubric design is used only during training. At inference time, RubricRM still generates the rubric and scores the candidates end to end. Further discussion of this design is provided in Appendix~\ref{sec:appendix_fixed_rubric}.

\paragraph{Dimension-Level Reward Design.}
When designing the reward, we consider two factors: first, whether the predicted score differences align in direction with the reference on each dimension, i.e., which of $A$ or $B$ is better; second, the magnitude of the score difference, as a large margin and a narrow margin should be rewarded differently even if both indicate the same preference. For each dimension $d_i$, let $(s_i^{A,gt},s_i^{B,gt})$ be the reference scores and $(s_i^{A,pred},s_i^{B,pred})$ be the rollout scores. We first compute the two score gaps:
\begin{equation}
\Delta_i^{gt}=s_i^{A,gt}-s_i^{B,gt},\qquad
\Delta_i^{pred}=s_i^{A,pred}-s_i^{B,pred}.
\end{equation}
The magnitude factor compares these gaps, normalized by the maximum possible gap difference:
\begin{equation}
b_i = 1-\frac{\left|\Delta_i^{pred}-\Delta_i^{gt}\right|}
{2(s_{\max}-s_{\min})}.
\end{equation}
We further assign a direction factor $\phi_i$ according to the signs of $\Delta_i^{gt}$ and $\Delta_i^{pred}$: $\phi_i=1.0$ when the signs match, $\phi_i=0.6$ when exactly one gap is zero, and $\phi_i=0.1$ when the two nonzero signs are opposite. This ordering reflects the relative severity of scoring errors: exact directional agreement is left unpenalized, tie mismatches are partially discounted, and complete reversals receive the largest penalty. The final rollout reward is the rubric-weighted aggregation:
\begin{equation}
R = \sum_{i=1}^{N} w_i b_i \phi_i.
\end{equation}
This reward prioritizes correct dimension-level preference direction through $\phi_i$, while $b_i$ further differentiates rollouts that have the same direction but different score-gap accuracy. Appendix~\ref{sec:appendix_reward_cases} gives concrete cases illustrating why the reward distinguishes reversed dimension directions, tie mismatches, and inaccurate score-difference magnitudes.

\paragraph{Saturated Group Filtering.}
Not all rollout groups provide reliable relative preference signals. For each conditioned input, GRPO samples $G$ responses and converts their rewards into group-relative advantages:
\begin{equation}
\hat{A}_i = \frac{R_i-\mu_R}{\sigma_R+\epsilon},
\end{equation}
where $\mu_R$ and $\sigma_R$ are the reward mean and standard deviation within the rollout group. This normalization is useful when rollouts differ meaningfully, but it becomes unreliable when all responses in a group receive nearly identical continuous rewards. In such low-spread groups, a tiny fluctuation in one rollout's reward can be amplified into a large relative advantage, causing the policy to update on noise rather than a meaningful scoring difference. We therefore filter low-standard-deviation groups by setting
\begin{equation}
A_i =
\begin{cases}
0, & \sigma_R < \sigma_{\text{th}},\\
\hat{A}_i, & \sigma_R \geq \sigma_{\text{th}}.
\end{cases}
\end{equation}
We use $\sigma_{\text{th}}=0.05$ for rewards in $[0,1]$. Because the decision is made from the reward spread of the current rollout group, the filtered groups change as the policy evolves; this is a dynamic training-time filter rather than an offline data removal step. Appendix~\ref{sec:appendix_saturation} gives the final GRPO objective, and Figure~\ref{fig:appendix_saturation_curves} reports the corresponding training-stability curves.

\section{Experiments}
This section evaluates RubricRM on pairwise preference benchmarks for text-to-image generation and image editing. We first describe the benchmarks, baselines, and implementation details, then compare RubricRM with specialized reward models and MLLM judges, and finally analyze the contribution of the two-stage training pipeline.
\begin{table}[h]
\centering
\caption{Comparison on text-to-image generation benchmarks. Bold and underlined scores denote the best and second-best results among reward models, respectively.}
\label{tab:benchmark}
\small
\setlength{\tabcolsep}{1.5pt}
\renewcommand{\arraystretch}{1.04}
\begin{tabularx}{\columnwidth}{@{}>{\raggedright\arraybackslash}p{0.49\columnwidth}>{\centering\arraybackslash}X>{\centering\arraybackslash}X>{\centering\arraybackslash}X@{}}
\toprule
\textbf{Model} & \textbf{MMRB2} & \makecell{\textbf{GenAI-}\\\textbf{Bench}} & \makecell{\textbf{GenAI-}\\\textbf{Bench-}\\\textbf{Verified}} \\
\midrule
\rowcolor{black!6}\multicolumn{4}{@{}l@{}}{\textit{Proprietary MLLMs}} \\
Claude Sonnet 4.6 & 70.8 & 65.8 & 75.3 \\
GPT-5.4 & 67.5 & 64.2 & 74.2 \\
Gemini 2.5 Pro & 70.5 & 67.8 & 77.4 \\
Gemini 3.1 Pro & 74.4 & 73.9 & 84.8 \\
\midrule
\rowcolor{black!6}\multicolumn{4}{@{}l@{}}{\textit{Open-source MLLMs}} \\
Qwen3-VL-8B & 61.2 & 63.3 & 72.5 \\
Qwen3-VL-235B-A22B & 66.6 & 61.5 & 69.7 \\
Qwen3.5-9B & 66.3 & 63.3 & 70.7 \\
Qwen3.5-397B-A17B & 72.7 & 66.2 & 77.0 \\
\midrule
\rowcolor{black!6}\multicolumn{4}{@{}l@{}}{\textit{Reward Models}} \\
HPSv2 & 55.0 & 68.8 & 78.1 \\
PickScore & 57.6 & 70.0 & 79.2 \\
HPSv3 & 60.2 & 70.9 & 81.0 \\
UnifiedReward-9B & 57.9 & 69.2 & 72.8 \\
UnifiedReward-Think-9B & 65.5 & 72.8 & 81.7 \\
UnifiedReward-Flex-8B & 69.2 & \underline{73.4} & \underline{84.2} \\
\rowcolor{green!6}RubricRM-Gen-4B (Ours) & \underline{70.5} & 73.2 & 83.1 \\
\rowcolor{green!6}RubricRM-Gen-9B (Ours) & \textbf{72.0} & \textbf{74.1} & \textbf{84.5} \\
\bottomrule
\end{tabularx}
\end{table}

\subsection{Experimental Setup}

\paragraph{Benchmarks}

\begin{table*}[t]
\centering
\caption{Comparison on image editing benchmarks. Avg denotes the benchmark-level average score. Bold and underlined scores denote the best and second-best results among reward models, respectively.}
\label{tab:benchmark_editing}
\footnotesize
\setlength{\tabcolsep}{2pt}
\renewcommand{\arraystretch}{1.08}
\begin{tabularx}{\textwidth}{@{}>{\raggedright\arraybackslash}p{0.25\textwidth}*{9}{>{\centering\arraybackslash}X}@{}}
\toprule
\multirow{2}{*}{\textbf{Model}} & \multirow{2}{*}{\textbf{MMRB2}} & \multicolumn{4}{c}{\textbf{EditReward-ERB}} & \multicolumn{4}{c}{\textbf{EditScore-ERB}} \\
\cmidrule(lr){3-6} \cmidrule(lr){7-10}
 &  & \textbf{K=2} & \textbf{K=3} & \textbf{K=4} & \textbf{Avg} & \textbf{PF} & \textbf{C} & \textbf{O} & \textbf{Avg} \\
\midrule
\rowcolor{black!6}\multicolumn{10}{@{}l@{}}{\textit{Proprietary MLLMs}} \\
Claude Sonnet 4.6 & 71.7 & 79.0 & 46.0 & 10.8 & 44.1 & 87.6 & 67.4 & 81.4 & 79.3 \\
GPT-5.4 & 68.5 & 78.2 & 40.7 & 12.2 & 42.5 & 83.7 & 62.8 & 76.3 & 74.6 \\
Gemini 2.5 Pro & 71.3 & 80.5 & 38.0 & 12.2 & 42.2 & 81.4 & 65.5 & 77.4 & 75.2 \\
Gemini 3.1 Pro & 74.9 & 78.2 & 45.3 & 14.9 & 45.0 & 87.7 & 71.6 & 84.1 & 81.6 \\
\midrule
\rowcolor{black!6}\multicolumn{10}{@{}l@{}}{\textit{Open-source MLLMs}} \\
Qwen3-VL-8B & 63.4 & 68.4 & 31.3 & 25.2 & 40.9 & 84.5 & 64.9 & 79.6 & 76.9 \\
Qwen3-VL-235B-A22B & 64.8 & 70.7 & 30.7 & 6.1 & 34.6 & 87.2 & 66.0 & 81.6 & 78.8 \\
Qwen3.5-9B & 64.4 & 75.2 & 33.3 & 7.4 & 37.4 & 79.5 & 60.6 & 74.6 & 72.0 \\
Qwen3.5-397B-A17B & 73.7 & 76.7 & 43.3 & 14.9 & 43.9 & 88.1 & 69.8 & 84.2 & 81.2 \\
\midrule
\rowcolor{black!6}\multicolumn{10}{@{}l@{}}{\textit{Reward Models}} \\
EditReward-7B & 67.2 & 56.5 & \underline{42.7} & 11.5 & 38.4 & 85.9 & 67.2 & 80.4 & 78.3 \\
EditScore-7B & 55.6 & 62.4 & 23.3 & 4.1 & 28.8 & 59.2 & 59.1 & 65.9 & 61.9 \\
\rowcolor{green!6}RubricRM-Edit-4B (Ours) & \underline{73.2} & \underline{76.7} & \textbf{48.0} & \underline{14.9} & \underline{45.5} & \textbf{88.7} & \underline{77.5} & \textbf{88.9} & \underline{85.5} \\
\rowcolor{green!6}RubricRM-Edit-9B (Ours) & \textbf{75.4} & \textbf{81.2} & 42.0 & \textbf{19.6} & \textbf{46.4} & \underline{88.6} & \textbf{79.7} & \underline{87.7} & \textbf{85.6} \\
\bottomrule
\end{tabularx}
\end{table*}

For text-to-image generation, we adopt the text-to-image subset of MMRB2 \cite{hu2025mmrb2}, GenAI-Bench \cite{jiang2024genai}, and our filtered GenAI-Bench-Verified. MMRB2 contains 1,000 pairwise comparison samples for assessing fine-grained preference judgment on text-to-image outputs. GenAI-Bench comprises approximately 1,700 pairwise comparisons and has been widely used in prior reward modeling work. We remove tie and both-bad cases, retaining only samples with a clear preference. Additionally, we observed that some samples in GenAI-Bench suffer from image mismatches or incorrect preference labels. To address this, we construct an expert-verified subset, GenAI-Bench-Verified, by inviting three expert annotators to re-label the candidate pairs and keeping only those samples where all three annotators agree and the consensus matches the original label. This filtering process yields a higher-confidence subset while maintaining compatibility with the original benchmark.

For image editing, we use the image editing subset of MMRB2 \cite{hu2025mmrb2}, EditReward-ERB \cite{wu2025editreward}, and EditScore-ERB \cite{luo2025editscore}. Since both \citet{wu2025editreward} and \citet{luo2025editscore} refer to their editing evaluation sets as EditReward-Bench, we use EditReward-ERB and EditScore-ERB to distinguish the two benchmark sources. MMRB2 also contains 1,000 test instances. EditReward-ERB evaluates K-way preference prediction with $K=2,3,4$ and also reports an overall benchmark score, which we denote as Avg in our table. EditScore-ERB contains 3,072 human-annotated preference pairs over 13 editing subtasks grouped into Subject, Appearance, Scene, and Advanced. It reports Prompt Following (PF), Consistency (C), Overall Quality (O), and an aggregate Avg score.

\paragraph{Baselines.}
We compare with general MLLM baselines, including open-source models Qwen3-VL-8B and Qwen3-VL-235B-A22B \cite{bai2025qwen3}, Qwen3.5-9B and Qwen3.5-397B-A17B \cite{qwen35blog}, as well as proprietary models Claude Sonnet 4.6 \cite{anthropic2026claudesonnet46}, GPT-5.4 \cite{openai2026gpt54}, Gemini 2.5 Pro \cite{google2025gemini25}, and Gemini 3.1 Pro \cite{google2026gemini31}. All MLLM judges are evaluated with chain-of-thought pairwise comparison prompts.
We also compare with specialized reward models. Text-to-image baselines include HPSv2 \cite{wu2023hpsv2}, PickScore \cite{kirstain2023pickapic}, HPSv3 \cite{ma2025hpsv3}, UnifiedReward-9B \cite{wang2025unified}, UnifiedReward-Think-9B \cite{wang2025think}, and UnifiedReward-Flex-8B \cite{wang2026flex}. For UnifiedReward and UnifiedReward-Think, we use their newly released Qwen3.5-9B checkpoints. Image editing baselines include EditReward-7B \cite{wu2025editreward} and EditScore-7B \cite{luo2025editscore}.

\paragraph{Implementation Details.}
We train separate RubricRM models for generation and editing on Qwen3.5-4B and Qwen3.5-9B backbones. The SFT stage uses learning rate $5\mathrm{e}{-6}$ with cosine decay and runs for 2 epochs. The GRPO stage uses learning rate $5\mathrm{e}{-7}$, group size $G=8$, and KL coefficient $\beta=0.05$. SFT and RL use disjoint training splits, and the RL split contains approximately 15K samples per task. Training is performed on 8 NVIDIA H20 GPUs. The task-specific inference prompt templates are shown in Appendices~\ref{sec:appendix_prompt_t2i} and~\ref{sec:appendix_prompt_edit}.

\subsection{Experimental Results}
Table~\ref{tab:benchmark} reports text-to-image results. RubricRM-Gen-9B achieves the highest scores among reward-model baselines on all three benchmarks, reaching 72.00 on MMRB2, 74.12 on GenAI-Bench, and 84.45 on GenAI-Bench-Verified. Compared with UnifiedReward-Flex-8B, the strongest prior reward-model baseline, RubricRM-Gen-9B improves by 2.80 points on MMRB2, 0.72 points on GenAI-Bench, and 0.25 points on GenAI-Bench-Verified. RubricRM-Gen-4B also performs competitively, obtaining 70.50, 73.20, and 83.11, despite using a smaller backbone than several general MLLM baselines.

Table~\ref{tab:benchmark_editing} reports image-editing results. RubricRM-Edit-9B achieves the highest reward-model scores on MMRB2 (75.40), EditReward-ERB Avg (46.40), and EditScore-ERB Avg (85.64). Among reward-model baselines, it also leads EditReward-ERB at K=2 and K=4, while RubricRM-Edit-4B obtains the highest K=3 score. The fine-grained EditScore-ERB metrics further support the motivation for dynamic rubrics. Image editing requires balancing instruction following, source-image consistency, and visual quality. RubricRM obtains high scores across all three aspects: RubricRM-Edit-4B achieves the highest Prompt Following (88.67) and Overall Quality (88.85), while RubricRM-Edit-9B achieves the highest Consistency (79.66) and Avg score. Appendix~\ref{sec:appendix_supp_results} provides the full category-wise results. This indicates that rubric-based scoring helps the model compare edited images through task-relevant criteria rather than relying on a single undifferentiated quality score.

\subsection{Ablation Study: Two-Stage Training}
Tables~\ref{tab:ablation_t2i} and~\ref{tab:ablation_edit} ablate the two stages of RubricRM by comparing the base Qwen3.5 backbones, the models after rubric-trajectory SFT, and the final models after dimension-level GRPO. The comparison is reported separately for text-to-image generation and image editing.
\providecommand{\ablgain}[1]{{\scriptsize\textcolor{green!55!black}{($\uparrow$#1)}}}

\begin{table}[h]
\centering
\caption{Text-to-image ablation of the two-stage training pipeline. Green deltas denote changes relative to the previous row within the same backbone.}
\label{tab:ablation_t2i}
\scriptsize
\setlength{\tabcolsep}{3pt}
\renewcommand{\arraystretch}{1.08}
\begin{tabular*}{\columnwidth}{@{\extracolsep{\fill}}lccc@{}}
\toprule
\textbf{Configuration} & \textbf{MMRB2} & \textbf{GenAI-Bench} & \textbf{GenAI-Bench-Verified} \\
\midrule
\textbf{Qwen3.5-4B} & 63.3 & 61.9 & 69.7 \\
\hspace{0.65em}+ SFT & 70.1 \ablgain{6.8} & 72.0 \ablgain{10.1} & 82.9 \ablgain{13.1} \\
\hspace{0.65em}+ RL  & 70.5 \ablgain{0.4} & 73.2 \ablgain{1.2}  & 83.1 \ablgain{0.2} \\
\midrule
\textbf{Qwen3.5-9B} & 66.9 & 63.4 & 72.5 \\
\hspace{0.65em}+ SFT & 70.3 \ablgain{3.4} & 73.0 \ablgain{9.6} & 83.2 \ablgain{10.7} \\
\hspace{0.65em}+ RL  & 72.0 \ablgain{1.7} & 74.1 \ablgain{1.2} & 84.5 \ablgain{1.2} \\
\bottomrule
\end{tabular*}
\end{table}

\begin{table}[h]
\centering
\caption{Image-editing ablation of the two-stage training pipeline. Green deltas denote changes relative to the previous row within the same backbone.}
\label{tab:ablation_edit}
\scriptsize
\setlength{\tabcolsep}{3pt}
\renewcommand{\arraystretch}{1.08}
\begin{tabular*}{\columnwidth}{@{\extracolsep{\fill}}lccc@{}}
\toprule
\textbf{Configuration} & \textbf{MMRB2} & \textbf{EditReward-ERB} & \textbf{EditScore-ERB} \\
\midrule
\textbf{Qwen3.5-4B} & 64.7 & 37.1 & 72.2 \\
\hspace{0.65em}+ SFT & 71.5 \ablgain{6.8} & 43.6 \ablgain{6.5} & 83.3 \ablgain{11.1} \\
\hspace{0.65em}+ RL  & 73.2 \ablgain{1.7} & 45.5 \ablgain{1.9} & 85.5 \ablgain{2.2} \\
\midrule
\textbf{Qwen3.5-9B} & 64.4 & 37.4 & 72.0 \\
\hspace{0.65em}+ SFT & 73.8 \ablgain{9.4} & 45.0 \ablgain{7.6} & 85.1 \ablgain{13.1} \\
\hspace{0.65em}+ RL  & 75.4 \ablgain{1.6} & 46.4 \ablgain{1.4} & 85.6 \ablgain{0.5} \\
\bottomrule
\end{tabular*}
\end{table}

Rubric-trajectory SFT accounts for most of the improvement. On text-to-image benchmarks, SFT improves the 4B backbone by 6.8 to 13.1 points and the 9B backbone by 3.4 to 10.7 points. On image-editing benchmarks, the corresponding gains are 6.5 to 11.1 points for 4B and 7.6 to 13.1 points for 9B. These gains suggest that a substantial part of the task lies in learning the rubric-based evaluation protocol itself, including criterion selection, weight assignment, and structured dimension-level scoring.

Dimension-level GRPO yields smaller but consistent additional gains across all benchmarks. This pattern is consistent with the role of the two stages: SFT establishes the rubric-scoring format, whereas GRPO calibrates its application by aligning predicted score gaps with the reference trajectory. Additional ablation results are provided in Appendix~\ref{sec:appendix_label_ablation}.

\section{Conclusion}

We introduce RubricRM, a generative reward model that represents visual evaluation as dynamic rubric generation followed by rubric-grounded scoring. This formulation makes the scoring protocol explicit: the model first determines what should be evaluated, then applies those criteria through weighted per-dimension scores. A two-stage training procedure separates these capabilities, using SFT for rubric formulation and GRPO for scoring alignment. Across generation and editing benchmarks, RubricRM improves over prior specialized reward models while preserving an inspectable judgment trace. 

\section*{Limitations}

RubricRM has several limitations. First, rubric trace synthesis relies on a proprietary teacher model, so teacher biases may be inherited by the student models. Second, the current framework evaluates static images and does not address video generation or temporal editing, where temporal consistency would require additional dimensions and scoring mechanisms. Future work will explore teacher-free rubric improvement, human-in-the-loop rubric correction, and extensions to video assessment.

\section*{Ethical Considerations}

RubricRM learns evaluative signals from human preference labels and teacher-synthesized rubric trajectories. These supervision sources may reflect subjective or systematic biases that can affect the learned rubrics and preference judgments. Although the explicit dimensions, weights, and scores improve inspectability, they should not be interpreted as objective standards or guaranteed causal explanations of a decision. This is especially important when the reward model is used to optimize downstream text-to-image generation models and image editing models, where biased reward signals may be amplified. We therefore recommend auditing RubricRM on the target content distribution and treating its rubric outputs as model judgments, not authoritative assessments. Since visual preference datasets may still contain people or sensitive content, users should adhere to the access conditions of the underlying datasets and consider content-specific audits before downstream deployment.

\bibliography{custom}

\appendix
\section{Method Details}
\label{sec:appendix_method}
This appendix provides additional methodological details that support the training pipeline: data coverage and taxonomy statistics, the advantage filtering rule used for GRPO, and representative dimension-level reward cases.
\begin{table}[h]
\centering
\caption{Second-level coverage of the text-to-image prompt taxonomy. Labels are multi-label; coverage is measured over all text-to-image training examples.}
\label{tab:t2i_secondary_labels}
\scriptsize
\setlength{\tabcolsep}{2.5pt}
\renewcommand{\arraystretch}{0.98}
\begin{tabularx}{\columnwidth}{@{}>{\raggedright\arraybackslash}p{0.32\columnwidth}>{\raggedright\arraybackslash}X r@{}}
\toprule
Primary & Secondary label & Cov. (\%) \\
\midrule
Common Objects & People & 45.81 \\
Style & Realistic Photography & 38.83 \\
Composition \& Perspective & Other & 26.34 \\
Common Objects & Everyday Items & 24.55 \\
Common Objects & Architecture \& Interior & 23.33 \\
Common Objects & Natural Scenery & 15.53 \\
Composition \& Perspective & Close-up \& Macro & 14.64 \\
Common Objects & Fictional Characters \& Creatures & 11.93 \\
Common Objects & Animals & 11.89 \\
Style & Illustration \& Hand-drawn & 11.86 \\
Common Objects & Plants \& Flowers & 8.41 \\
Style & 3D Rendering & 7.66 \\
Logical Reasoning & Spatial Relations & 7.41 \\
World Knowledge & History \& Culture & 7.24 \\
Common Objects & Vehicles & 7.03 \\
Style & Traditional Art & 5.55 \\
Composition \& Perspective & Panoramic \& Wide-angle & 5.46 \\
Style & Abstract Art & 5.06 \\
Text Rendering & English & 4.29 \\
Composition \& Perspective & Overhead \& Low-angle & 4.24 \\
Style & Cyberpunk \& Steampunk & 4.12 \\
Style & Anime & 4.04 \\
Common Objects & Food \& Drinks & 3.97 \\
World Knowledge & IP Characters \& Brands & 3.96 \\
Logical Reasoning & Counting & 3.94 \\
World Knowledge & Geography \& Landmarks & 3.24 \\
Style & Minimalist \& Flat & 2.78 \\
World Knowledge & Famous People & 2.48 \\
Composition \& Perspective & Symmetrical & 2.39 \\
Logical Reasoning & Common Sense & 1.62 \\
Style & Pixel Art & 1.36 \\
Text Rendering & Logo & 1.26 \\
World Knowledge & Professional Domains & 1.11 \\
Logical Reasoning & Causal \& Temporal & 1.05 \\
Text Rendering & Multilingual \& Symbols & 0.90 \\
Text Rendering & Handwriting \& Calligraphy & 0.68 \\
Logical Reasoning & Linguistic Logic & 0.64 \\
Text Rendering & Chinese & 0.60 \\
Logical Reasoning & Scientific Reasoning & 0.59 \\
Style & Pop Art & 0.50 \\
Style & Ink Wash Painting & 0.30 \\
\bottomrule
\end{tabularx}
\end{table}

\begin{table}[h]
\centering
\caption{Second-level coverage of the image-editing instruction taxonomy. Labels are multi-label; coverage is measured over all image-editing training examples.}
\label{tab:edit_secondary_labels}
\scriptsize
\setlength{\tabcolsep}{2.5pt}
\renewcommand{\arraystretch}{0.98}
\begin{tabularx}{\columnwidth}{@{}>{\raggedright\arraybackslash}p{0.34\columnwidth}>{\raggedright\arraybackslash}X r@{}}
\toprule
Primary & Secondary label & Cov. (\%) \\
\midrule
Object Editing & Add Object & 21.01 \\
Attribute Editing & Color Change & 14.30 \\
Object Editing & Remove Object & 12.52 \\
Background Editing & Background Replacement & 11.82 \\
Text Editing & Text Editing & 11.17 \\
Object Editing & Replace Object & 10.77 \\
Style Editing & Art Style Transfer & 8.01 \\
Attribute Editing & Material \& Texture & 5.75 \\
Attribute Editing & Appearance \& Shape & 5.59 \\
Reasoning-based Editing & Implicit Intent & 4.68 \\
Reasoning-based Editing & Complex Instruction & 4.44 \\
Background Editing & Lighting \& Environment & 4.06 \\
Inpainting \& Repair & Watermark \& Occlusion Removal & 3.65 \\
Background Editing & Season \& Weather Change & 3.05 \\
Inpainting \& Repair & Blemish Removal & 2.85 \\
Object Editing & Move \& Rearrange & 2.63 \\
Pose \& Action Editing & Body Pose & 2.21 \\
Attribute Editing & Size Adjustment & 1.25 \\
Style Editing & Era Style & 1.08 \\
Attribute Editing & Quantity Change & 0.98 \\
Style Editing & Material Style & 0.96 \\
Face Editing & Facial Attributes & 0.95 \\
Style Editing & Tone \& Filter & 0.64 \\
Pose \& Action Editing & Expression Change & 0.56 \\
Background Editing & Background Removal & 0.51 \\
Viewpoint Change & Viewpoint Change & 0.30 \\
Multi-image Composition & Multi-image Composition & 0.27 \\
Quality Enhancement & Denoise \& Deblur & 0.14 \\
Quality Enhancement & Super Resolution & 0.12 \\
Face Editing & Age Change & 0.11 \\
\bottomrule
\end{tabularx}
\end{table}

\vspace{-0.2cm}
\subsection{Training Data and Taxonomy Statistics}
\label{sec:appendix_data}
\paragraph{RL split construction.}
The RL split is selected with stratified sampling over taxonomy annotations. We start from a sampled split and iteratively swap examples when a secondary label deviates from the target coverage. This keeps the maximum secondary-label coverage drift below 0.5 percentage points for both tasks, preserving low-frequency categories while reducing overrepresentation of frequent categories.

\begin{figure*}[h]
\centering
\begin{minipage}[t]{0.48\textwidth}
\centering
\includegraphics[width=\linewidth]{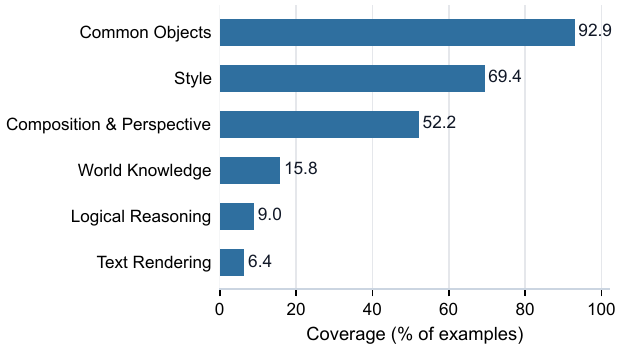}
\vspace{1pt}
{\small\textbf{Text-to-image generation}\par}
\end{minipage}
\hfill
\begin{minipage}[t]{0.48\textwidth}
\centering
\includegraphics[width=\linewidth]{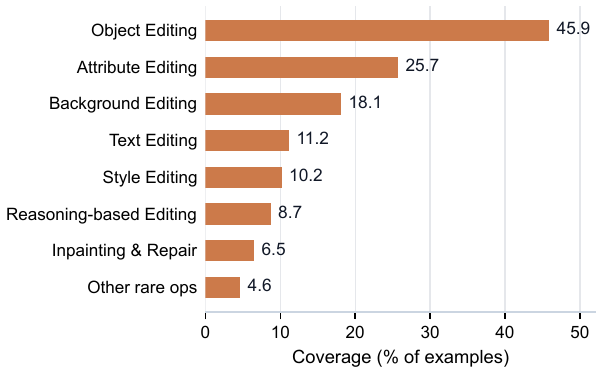}
\vspace{1pt}
{\small\textbf{Image editing}\par}
\end{minipage}
\vspace{-2pt}
\caption{Primary-category coverage of the rubric-learning data for text-to-image generation and image editing.}
\label{fig:appendix_primary_coverage}
\end{figure*}

\paragraph{Primary-category coverage.}
For text-to-image generation, the most frequent primary categories are Common Objects (92.4\%), Style (69.0\%), and Composition \& Perspective (52.1\%). Long-tail but important categories remain represented: World Knowledge covers 15.8\%, Logical Reasoning covers 9.0\%, and Text Rendering covers 6.5\% of examples. For image editing, Object Editing covers 45.9\%, Attribute Editing 25.7\%, Background Editing 18.2\%, Text Editing 11.2\%, Style Editing 10.2\%, Reasoning-based Editing 8.7\%, and Inpainting \& Repair 6.5\%.
Figure~\ref{fig:appendix_primary_coverage} visualizes these primary-category distributions for both tasks.
\begin{table*}[h]
\centering
\caption{Semantic aggregation of parsed text-to-image rubric dimensions. Dim. share is computed over parsed teacher-generated rubric dimensions; avg. wt. is the average teacher-assigned dimension weight.}
\label{tab:t2i_rubric_atoms}
\scriptsize
\setlength{\tabcolsep}{3.2pt}
\renewcommand{\arraystretch}{1.04}
\begin{tabularx}{\textwidth}{@{}llrrX@{}}
\toprule
Semantic group & Rubric atom & Dim. share (\%) & Avg. wt. & Representative original dimension titles \\
\midrule
Semantic Alignment & Prompt alignment & 7.99 & 32.7 & Content Accuracy; Semantic Accuracy; Semantic Alignment \\
Semantic Alignment & Text rendering & 1.93 & 32.2 & Text Accuracy; Text Rendering; Text Rendering Accuracy \\
Semantic Alignment & Reasoning \& knowledge & 1.78 & 29.7 & Composition \& Spatial Logic; Logical Consistency; Compositional Logic \\
Subject \& Attributes & Subject fidelity & 14.89 & 30.3 & Subject Accuracy; Subject Representation; Character Accuracy \\
Subject \& Attributes & Anatomy / pose & 9.58 & 27.9 & Anatomical Integrity; Anatomical Correctness; Anatomical Coherence \\
Subject \& Attributes & Attribute fidelity & 8.64 & 26.5 & Color Palette; Attire Accuracy; Lighting \& Color Palette \\
Scene \& Composition & Scene / background & 11.81 & 24.3 & Environmental Context; Setting \& Atmosphere; Environment \& Lighting \\
Scene \& Composition & Composition / perspective & 7.25 & 25.6 & Spatial Composition; Composition \& Perspective; Composition \& Framing \\
Scene \& Composition & Lighting / atmosphere & 6.01 & 24.2 & Lighting \& Atmosphere; Atmosphere \& Lighting; Lighting \& Contrast \\
Style \& Quality & Visual quality & 17.94 & 26.2 & Image Quality \& Coherence; Structural Coherence; Image Quality \\
Style \& Quality & Style adherence & 9.25 & 26.8 & Stylistic Execution; Artistic Style; Image Quality \& Aesthetics \\
Style \& Quality & Realism & 2.94 & 24.2 & Image Quality \& Realism; Visual Quality \& Realism; Realism \& Image Quality \\
\bottomrule
\end{tabularx}
\end{table*}

\begin{table*}[h]
\centering
\caption{Semantic aggregation of parsed image-editing rubric dimensions. Dim. share is computed over parsed teacher-generated rubric dimensions; avg. wt. is the average teacher-assigned dimension weight.}
\label{tab:edit_rubric_atoms}
\scriptsize
\setlength{\tabcolsep}{3.2pt}
\renewcommand{\arraystretch}{1.04}
\begin{tabularx}{\textwidth}{@{}llrrX@{}}
\toprule
Semantic group & Rubric atom & Dim. share (\%) & Avg. wt. & Representative original dimension titles \\
\midrule
Instruction \& Edit & Instruction alignment & 8.88 & 39.6 & Instruction Alignment; Instruction Compliance; Instruction Adherence \\
Instruction \& Edit & Object edit & 7.72 & 36.4 & Object Removal; Object Addition; Object Replacement \\
Instruction \& Edit & Text editing & 5.47 & 35.9 & Text Accuracy; Text Rendering Accuracy; Text Placement \\
Instruction \& Edit & Attribute edit & 4.45 & 35.4 & Color Modification; Material Transformation; Color Transformation \\
Preservation & Source preservation & 20.90 & 31.4 & Content Preservation; Image Preservation; Structural Preservation \\
Preservation & Subject preservation & 5.78 & 33.5 & Subject Preservation; Context Preservation; Foreground Preservation \\
Scene Integration & Background edit & 10.85 & 30.9 & Background Preservation; Inpainting Quality; Background Reconstruction \\
Scene Integration & Visual integration & 9.59 & 27.7 & Realism \& Integration; Visual Integration; Realism \& Blending \\
Scene Integration & Spatial placement & 2.51 & 31.3 & Spatial Placement; Spatial Positioning; Placement Accuracy \\
Quality \& Realism & Quality / artifacts & 13.12 & 31.8 & Image Quality; Visual Quality; Editing Accuracy \\
Quality \& Realism & Realism & 6.43 & 26.2 & Visual Realism; Image Quality \& Realism; Image Realism \\
Quality \& Realism & Style transfer & 3.97 & 35.5 & Stylistic Transformation; Stylistic Accuracy; Aesthetic Quality \\
Quality \& Realism & Reasoning & 0.32 & 38.4 & Semantic Accuracy; Semantic Transformation; Semantic Preservation \\
\bottomrule
\end{tabularx}
\end{table*}

\paragraph{Secondary labels and co-occurrence.}
The most frequent text-to-image secondary labels are People (45.8\%), Realistic Photography (38.8\%), Other Composition (26.3\%), Everyday Items (24.6\%), and Architecture \& Interior (23.3\%). The strongest primary-category co-occurrences are Common Objects+Style (63.6\%), Common Objects+Composition \& Perspective (49.5\%), and Composition \& Perspective+Style (37.3\%). These co-occurrences show that many prompts simultaneously specify subject matter, visual style, and composition, motivating instruction-specific rubric dimensions.

For image editing, the most frequent secondary labels are Add Object (21.0\%), Color Change (14.3\%), Remove Object (12.5\%), Background Replacement (11.8\%), Text Editing (11.2\%), Replace Object (10.8\%), and Art Style Transfer (8.0\%). The strongest co-occurrences are Attribute Editing+Object Editing (6.6\%), Object Editing+Reasoning-based Editing (5.2\%), Inpainting \& Repair+Object Editing (4.5\%), and Background Editing+Object Editing (3.4\%).
Tables~\ref{tab:t2i_secondary_labels} and~\ref{tab:edit_secondary_labels} list the complete second-level taxonomy coverage used to diagnose long-tail categories and construct balanced training splits.

\paragraph{Rubric-dimension aggregation.}
To check whether synthesized rubrics cover task-relevant evaluation aspects, we parse teacher-generated dimension titles and aggregate them into semantic groups. Tables~\ref{tab:t2i_rubric_atoms} and~\ref{tab:edit_rubric_atoms} show the resulting dimension shares and average teacher-assigned weights. The aggregated dimensions align with the two task settings: text-to-image rubrics emphasize subject fidelity, visual quality, scene composition, style, and prompt alignment, whereas image-editing rubrics additionally emphasize source preservation and edit integration.
\subsection{Fixed-Rubric Rollout during GRPO}
\label{sec:appendix_fixed_rubric}
Our dimension-level reward compares the predicted and reference score gaps for each evaluation dimension, which requires a one-to-one correspondence between the rollout and reference dimensions. If each rollout generated its own dimensions and weights, different rollouts would be evaluated under different criteria, making the dimension-level rewards no longer directly comparable and introducing substantial training noise. We therefore treat the teacher-generated rubric as a training-time conditioning context and roll out only the subsequent dimension-level assessments and scores. This design provides a denser signal than rewarding only the final binary preference. Importantly, the restriction applies only during GRPO: rubric generation is learned from complete trajectories during SFT, and at inference time no reference rubric is provided, and the model generates an input-specific rubric and scores the candidates end to end. Thus, GRPO should be understood as calibrating rubric-conditioned scoring rather than directly optimizing rubric generation.

\subsection{GRPO Objective and Saturated-Group Filtering}
\label{sec:appendix_saturation}

Section~\ref{sec:method} defines the dimension-level reward, group-relative advantage normalization, and saturated-group filtering rule. Using the filtered advantages $A_i$, the final GRPO objective is
\begin{equation}
\begin{aligned}
\mathcal{J}_{\mathrm{GRPO}}(\theta)
&= \mathbb{E}_{\tilde{x}, \{o_i\}}
\Bigg[
\frac{1}{G} \sum_{i=1}^{G}
\Big(
\min\!\big(
\rho_i A_i, \\
&\quad
\mathrm{clip}(\rho_i, 1-\epsilon, 1+\epsilon) A_i
\big)
- \beta D_{\mathrm{KL}}^{(i)}
\Big)
\Bigg],
\end{aligned}
\end{equation}
where $\rho_i = \pi_\theta(o_i \mid \tilde{x}) / \pi_{\theta_{\mathrm{old}}}(o_i \mid \tilde{x})$, and $D_{\mathrm{KL}}^{(i)}$ estimates the KL divergence between the response $o_i$ and a reference policy. The empirical training-stability curves for saturated-group filtering are reported in Figure~\ref{fig:appendix_saturation_curves}.

\begin{table*}[h]
\centering
\footnotesize
\setlength{\tabcolsep}{4pt}
\renewcommand{\arraystretch}{0.92}
\begin{adjustbox}{max width=\textwidth}
\begin{tabular}{cllccccccc}
\toprule
Case & Rollout & Dimension & $w_i$ & Reference A/B & Predicted A/B & $b_i$ & $\phi_i$ & $r_i$ & Total $R$ \\
\midrule
\multirow{3}{*}{\makecell{1\\Reverse}} & \multirow{3}{*}{--} & Camera Perspective & 0.40 & 1 / 4 & 0 / 4 & 0.875 & 1.0 & 0.875 & \multirow{3}{*}{0.6313} \\
& & Thematic Depiction & 0.30 & 4 / 2 & 2 / 3 & 0.625 & 0.1 & 0.0625 & \\
& & Structural Integrity & 0.30 & 0 / 1 & 0 / 2 & 0.875 & 1.0 & 0.875 & \\
\midrule
\multirow{6}{*}{\makecell{2\\Tie}} & \multirow{3}{*}{1} & Prompt Adherence & 0.40 & 2 / 1 & 3 / 1 & 0.875 & 1.0 & 0.875 & \multirow{3}{*}{0.950} \\
& & Anatomical Correctness & 0.40 & 0 / 0 & 0 / 0 & 1.0 & 1.0 & 1.0 & \\
& & Subject Separation & 0.20 & 2 / 0 & 2 / 0 & 1.0 & 1.0 & 1.0 & \\
\cmidrule(lr){2-10}
& \multirow{3}{*}{2} & Prompt Adherence & 0.40 & 2 / 1 & 3 / 1 & 0.875 & 1.0 & 0.875 & \multirow{3}{*}{0.760} \\
& & Anatomical Correctness & 0.40 & 0 / 0 & 1 / 0 & 0.875 & 0.6 & 0.525 & \\
& & Subject Separation & 0.20 & 2 / 0 & 2 / 0 & 1.0 & 1.0 & 1.0 & \\
\midrule
\multirow{3}{*}{\makecell{3\\Margin}} & \multirow{3}{*}{--} & Entity Accuracy & 0.40 & 4 / 0 & 4 / 0 & 1.0 & 1.0 & 1.0 & \multirow{3}{*}{0.8875} \\
& & Spatial Obstruction & 0.30 & 4 / 0 & 1 / 0 & 0.625 & 1.0 & 0.625 & \\
& & Relative Size & 0.30 & 4 / 0 & 4 / 0 & 1.0 & 1.0 & 1.0 & \\
\bottomrule
\end{tabular}
\end{adjustbox}
\caption{Dimension-level reward cases illustrating a reversed dimension direction, a tie mismatch, and an underestimated score-difference margin. Here $r_i=b_i\phi_i$ denotes the per-dimension reward.}
\label{tab:reward_cases}
\end{table*}
\subsection{Dimension-Level Reward Case Studies}
\label{sec:appendix_reward_cases}

Dimension-level rewards supervise more than the final pairwise preference. The following training-rollout cases illustrate the three behaviors encoded by our reward: reversed dimension directions receive the strongest discount, tie mismatches receive a partial discount, and score-difference magnitudes are still rewarded when their directions are correct. Table~\ref{tab:reward_cases} reports the per-dimension terms and the final weighted reward for each case.

\paragraph{Case 1: Correct final preference with a reversed dimension.}
For the prompt ``writer creates stories, Overhead shot,'' both the reference and the rollout select candidate $B$ as the final preference. However, Table~\ref{tab:reward_cases} shows that the rollout reverses the reference direction on \textit{Thematic Depiction}: the reference scores indicate $A>B$, whereas the predicted scores indicate $A<B$. A final-preference-only reward would not expose this incorrect scoring rationale.

\paragraph{Case 2: Tie mismatches receive a partial discount.}
For the prompt ``Person lying down with legs up, next to a dog,'' two rollouts both select the correct final preference $A$. Rollout 1 in Table~\ref{tab:reward_cases} preserves the reference tie on \textit{Anatomical Correctness}, whereas Rollout 2 introduces an unsupported $A>B$ direction. This tie mismatch applies $\lambda_{\text{tie}}=0.6$ and lowers the aggregate reward from $0.950$ to $0.760$.

\paragraph{Case 3: Correct direction with an inaccurate margin.}
For the prompt ``a Speaker is partially obstructed by a Snail, and the Speaker is larger than the Snail,'' the rollout selects the correct final preference $A$ and matches the direction on every dimension. However, Table~\ref{tab:reward_cases} shows that it underestimates the reference margin on \textit{Spatial Obstruction}, predicting $1/0$ instead of the reference $4/0$.

\section{Experiments}
\label{sec:appendix_experiments}

\subsection{Supplementary Results}
\paragraph{Detailed editing results.}
\begin{table*}[htbp]
\centering
\caption{Detailed results on EditScore-ERB across four task categories and overall. PF = Prompt Following, C = Consistency, O = Overall Quality. Scores are percentages. Bold and underlined scores denote the best and second-best results among reward models, respectively.}
\label{tab:esbench_detail}
\setlength{\tabcolsep}{1.5pt}
\renewcommand{\arraystretch}{1.08}
\scriptsize
\begin{tabularx}{\textwidth}{@{}>{\raggedright\arraybackslash}p{0.18\textwidth}*{15}{>{\centering\arraybackslash}X}@{}}
\toprule
\multirow{2}{*}{\textbf{Model}} & \multicolumn{3}{c}{\textbf{Subject}} & \multicolumn{3}{c}{\textbf{Appearance}} & \multicolumn{3}{c}{\textbf{Scene}} & \multicolumn{3}{c}{\textbf{Advanced}} & \multicolumn{3}{c}{\textbf{Overall}} \\
\cmidrule(lr){2-4} \cmidrule(lr){5-7} \cmidrule(lr){8-10} \cmidrule(lr){11-13} \cmidrule(lr){14-16}
 & PF & C & O & PF & C & O & PF & C & O & PF & C & O & PF & C & O \\
\midrule
\rowcolor{black!6}\multicolumn{16}{@{}l@{}}{\textit{\textbf{Proprietary MLLMs}}} \\
Claude Sonnet 4.6 & 83.9 & 64.5 & 82.9 & 86.8 & 69.2 & 76.2 & 93.8 & 74.5 & 88.8 & 88.7 & 63.5 & 83.2 & 87.6 & 67.4 & 81.4 \\
GPT-5.4 & 86.1 & 56.2 & 79.7 & 82.8 & 62.8 & 73.3 & 91.9 & 72.4 & 82.5 & 79.2 & 62.8 & 74.8 & 83.7 & 62.8 & 76.3 \\
Gemini 2.5 Pro & 81.6 & 61.3 & 77.5 & 75.9 & 69.8 & 75.9 & 82.6 & 72.4 & 80.1 & 85.7 & 62.3 & 79.7 & 81.4 & 65.5 & 77.4 \\
Gemini 3.1 Pro & 86.5 & 65.9 & 86.3 & 83.0 & 75.1 & 77.8 & 90.1 & 76.4 & 90.8 & 91.7 & 70.3 & 87.5 & 87.7 & 71.6 & 84.1 \\
\midrule
\rowcolor{black!6}\multicolumn{16}{@{}l@{}}{\textit{\textbf{Open-source MLLMs}}} \\
Qwen3-VL-8B & 85.4 & 67.6 & 84.7 & 84.8 & 64.8 & 75.2 & 88.4 & 74.6 & 82.2 & 81.5 & 57.5 & 78.9 & 84.5 & 64.9 & 79.6 \\
Qwen3-VL-235B-A22B & 90.0 & 69.9 & 85.4 & 85.8 & 62.5 & 75.9 & 91.2 & 76.4 & 87.2 & 84.7 & 62.3 & 82.9 & 87.2 & 66.0 & 81.6 \\
Qwen3.5-9B & 79.8 & 55.7 & 74.1 & 79.3 & 61.3 & 71.4 & 87.5 & 72.4 & 84.0 & 77.8 & 58.3 & 75.8 & 79.5 & 60.6 & 74.6 \\
Qwen3.5-397B-A17B & 92.2 & 71.9 & 86.5 & 84.7 & 69.9 & 78.9 & 90.1 & 77.5 & 91.2 & 86.8 & 64.0 & 85.8 & 88.1 & 69.8 & 84.2 \\
\midrule
\rowcolor{black!6}\multicolumn{16}{@{}l@{}}{\textit{\textbf{Reward Models}}} \\
EditReward-7B & 86.6 & 66.2 & 83.1 & 84.2 & 65.8 & 76.5 & \textbf{96.2} & 75.5 & 89.2 & 83.2 & 65.3 & 79.9 & 85.9 & 67.2 & 80.4 \\
EditScore-7B & 59.0 & 58.5 & 74.0 & 57.3 & 61.2 & 66.3 & 74.4 & 73.5 & 78.9 & 53.6 & 50.3 & 52.9 & 59.2 & 59.1 & 65.9 \\
\rowcolor{green!6}RubricRM-Edit-4B (Ours) & \textbf{90.7} & \underline{77.8} & \textbf{91.5} & \textbf{86.4} & \underline{83.3} & \textbf{87.5} & 90.7 & \underline{80.4} & \underline{90.1} & \textbf{88.7} & \underline{70.5} & \textbf{88.0} & \textbf{88.7} & \underline{77.5} & \textbf{88.9} \\
\rowcolor{green!6}RubricRM-Edit-9B (Ours) & \underline{90.6} & \textbf{78.3} & \underline{90.7} & \underline{86.0} & \textbf{84.0} & \underline{86.2} & \underline{93.0} & \textbf{85.9} & \textbf{90.8} & \underline{87.7} & \textbf{73.6} & \underline{85.4} & \underline{88.6} & \textbf{79.7} & \underline{87.7} \\
\bottomrule
\end{tabularx}
\end{table*}

\begin{figure*}[h]
\centering
\setlength{\tabcolsep}{4pt}
\begin{tabular}{ccc}
\includegraphics[width=0.25\textwidth]{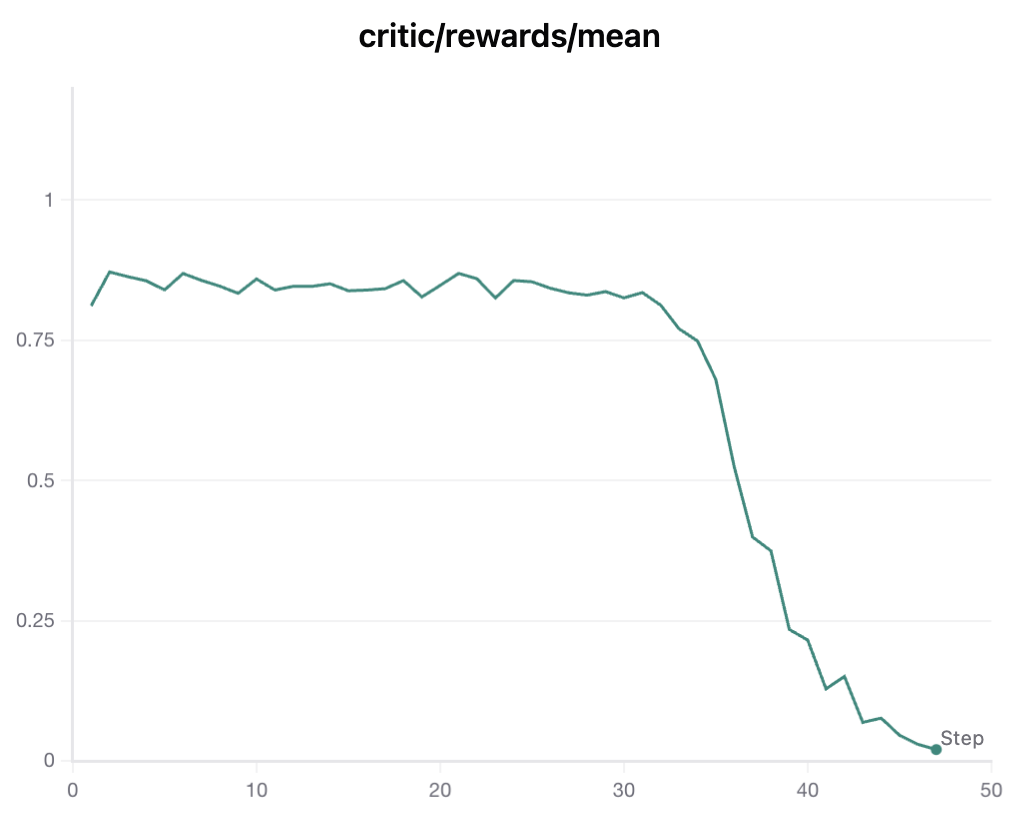} &
\includegraphics[width=0.25\textwidth]{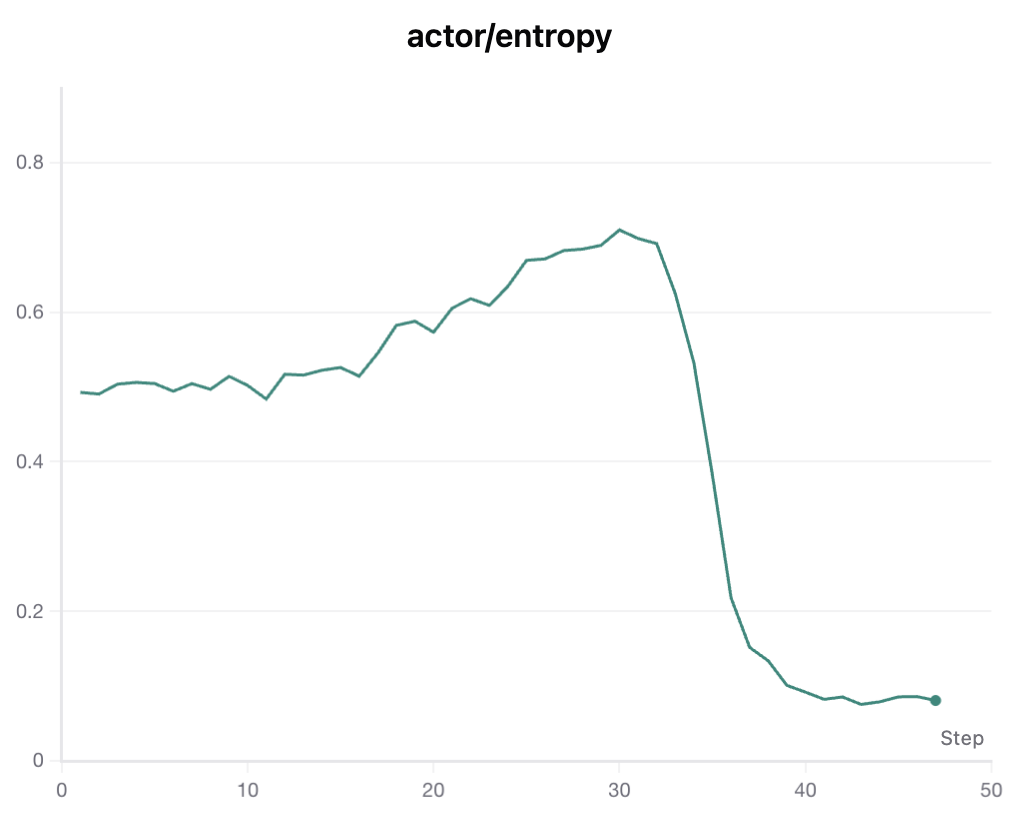} &
\includegraphics[width=0.25\textwidth]{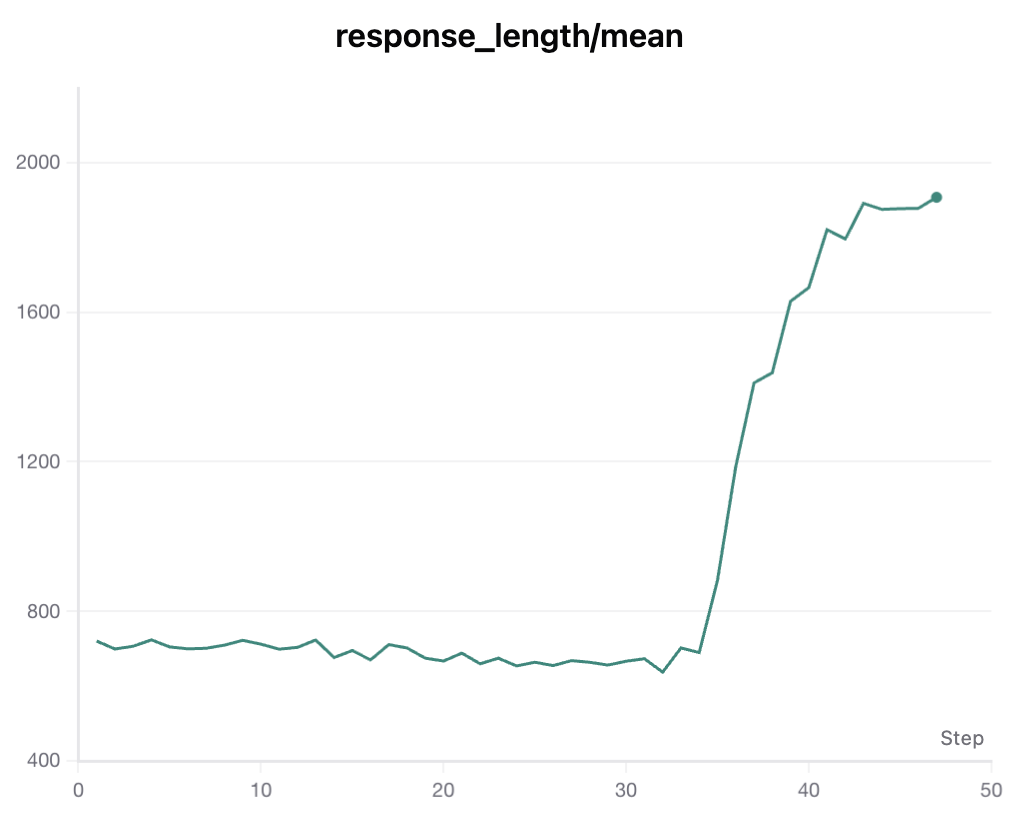} \\[3pt]
\includegraphics[width=0.25\textwidth]{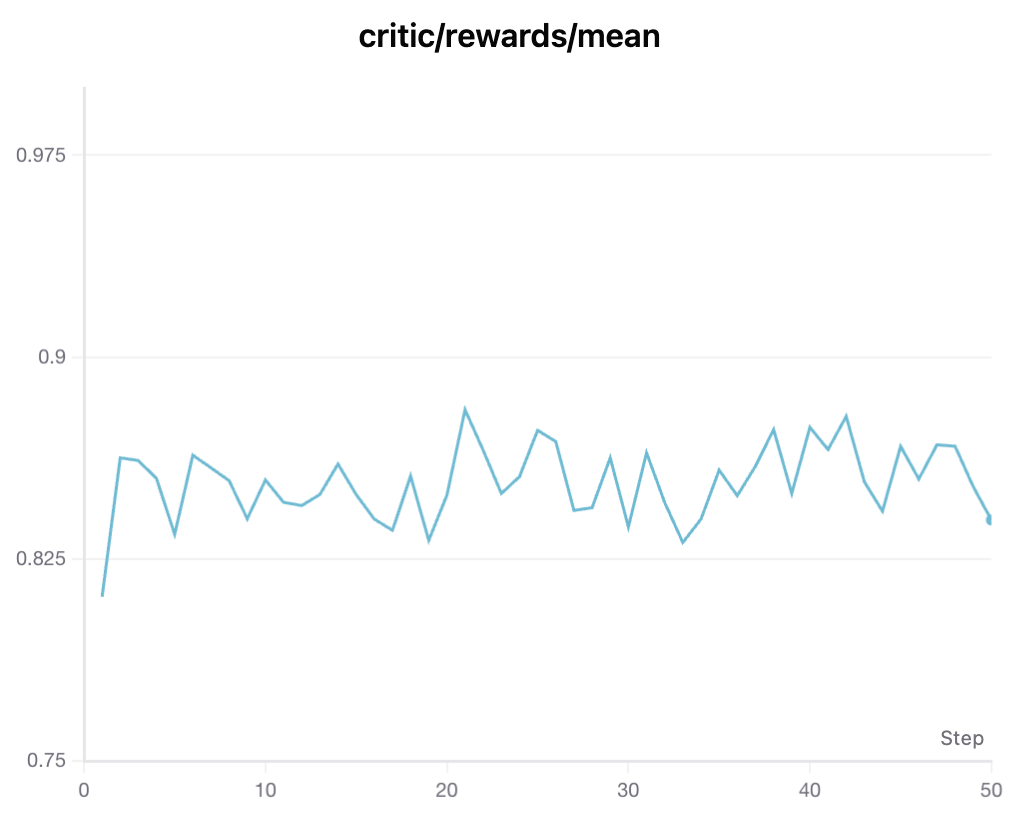} &
\includegraphics[width=0.25\textwidth]{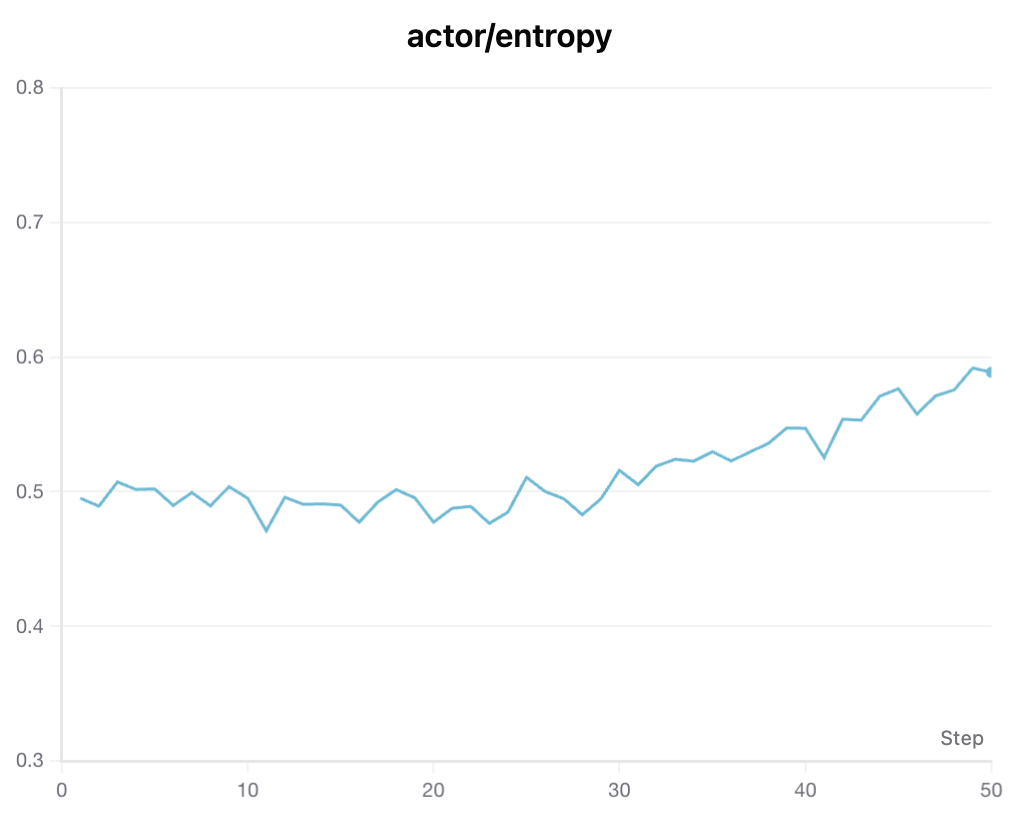} &
\includegraphics[width=0.25\textwidth]{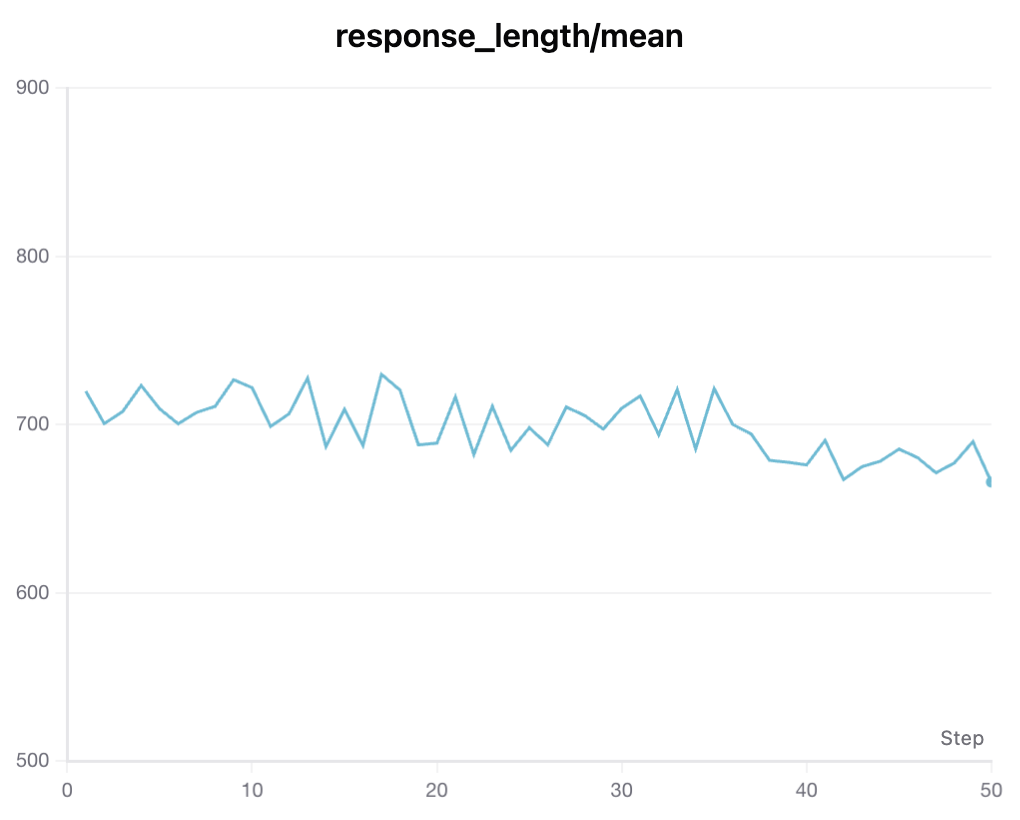}
\end{tabular}
\caption{Effect of saturated-group filtering on GRPO training stability. The columns show reward mean, entropy, and response length from left to right; the top row is without filtering and the bottom row is with filtering. Without filtering, late training shows reward-mean collapse, sharp response-length growth, and entropy drift followed by collapse. Filtering low-standard-deviation groups keeps all three signals stable.}
\label{fig:appendix_saturation_curves}
\end{figure*}

\begin{table*}[h]
\centering
\caption{Label-only versus rubric-based SFT ablation. Scores are percentages; deltas in Label-only SFT rows are relative to Rubric SFT within each backbone.}
\label{tab:label_ablation}
\newcommand{\labelgain}[1]{{\scriptsize\textcolor{green!55!black}{($\uparrow$#1)}}}
\newcommand{\labelloss}[1]{{\scriptsize\textcolor{red!70!black}{($\downarrow$#1)}}}
\scriptsize
\setlength{\tabcolsep}{3pt}
\renewcommand{\arraystretch}{1.12}
\begin{tabularx}{\textwidth}{@{}>{\raggedright\arraybackslash}p{0.14\textwidth}>{\raggedright\arraybackslash}p{0.16\textwidth}*{6}{>{\raggedright\arraybackslash}X}@{}}
\toprule
\multirow{2}{*}{\textbf{Backbone}} &
\multirow{2}{*}{\textbf{Supervision}} &
\multicolumn{3}{c}{\textbf{Text-to-Image}} &
\multicolumn{3}{c}{\textbf{Image Editing}} \\
\cmidrule(lr){3-5} \cmidrule(lr){6-8}
& & \textbf{MMRB2} & \makecell{\textbf{GenAI-}\\\textbf{Bench}} & \makecell{\textbf{GenAI-}\\\textbf{Bench-}\\\textbf{Verified}} & \textbf{MMRB2} & \makecell{\textbf{EditReward-}\\\textbf{ERB}} & \makecell{\textbf{EditScore-}\\\textbf{ERB}} \\
\midrule
\multirow{2}{*}{Qwen3.5-4B} & Rubric SFT & 70.1 & 72.0 & 82.9 & 71.5 & 43.6 & 83.3 \\
& Label-only SFT & 65.1 \labelloss{5.0} & 68.7 \labelloss{3.3} & 77.8 \labelloss{5.1} & 69.2 \labelloss{2.3} & 39.4 \labelloss{4.2} & 85.4 \labelgain{2.1} \\
\midrule
\multirow{2}{*}{Qwen3.5-9B} & Rubric SFT & 70.3 & 73.0 & 83.2 & 73.8 & 45.0 & 85.1 \\
& Label-only SFT & 66.4 \labelloss{3.9} & 71.1 \labelloss{1.9} & 80.9 \labelloss{2.3} & 69.2 \labelloss{4.6} & 40.8 \labelloss{4.2} & 83.9 \labelloss{1.2} \\
\bottomrule
\end{tabularx}
\end{table*}

Table~\ref{tab:esbench_detail} expands the image-editing results on EditScore-ERB by reporting Prompt Following, Consistency, and Overall Quality for each task category. These detailed scores support the aggregate trend in Table~\ref{tab:benchmark_editing}: RubricRM remains competitive across subject, appearance, scene, and advanced editing categories rather than improving only a single aspect.
\label{sec:appendix_supp_results}

\paragraph{Label-only supervision ablation.}
\label{sec:appendix_label_ablation}

Table~\ref{tab:label_ablation} compares direct label supervision with rubric-based SFT. This ablation uses the same backbones and preference data, but replaces the teacher-synthesized rubric trajectory with a direct pairwise preference target. It therefore isolates whether the improvement comes only from the training data and preference labels, or also from the rubric-based scoring paradigm.

The label-only models improve over the base models, showing that the preference labels are useful on their own. However, they remain consistently below rubric-based SFT on most benchmarks. For text-to-image generation, the gap is especially clear: label-only SFT is lower than rubric-based SFT by 3.3--5.1 points for the 4B backbone and 1.9--3.9 points for the 9B backbone. On image editing, label-only SFT also falls behind on MMRB2 and EditReward-ERB, where instruction following and edit-specific criteria are central. These results indicate that RubricRM's gains are not merely due to additional exposure to the same training examples. They also come from the proposed paradigm: the rubric trajectory teaches the model to decompose an instruction into evaluation criteria before making the pairwise judgment.

\paragraph{Saturated-group filtering analysis.}
Figure~\ref{fig:appendix_saturation_curves} provides an empirical view of the saturated-group filtering mechanism described in Section~\ref{sec:method}. In our GRPO stage, rewards are continuous dimension-level scores and advantages are normalized within each rollout group. When all rollouts for the same input receive nearly identical rewards, the group contains little reliable relative preference signal; nevertheless, within-group normalization can amplify small numerical fluctuations into large policy-gradient updates. This is reflected in the unfiltered curves: the reward mean remains high in early training but collapses sharply in the late stage, response length grows abruptly, and entropy first drifts upward before collapsing. These coupled changes indicate that low-discrimination rollout groups can destabilize the policy update rather than improve rubric application. After filtering groups whose reward standard deviation falls below the threshold, the reward mean fluctuates within a narrow range, response length remains bounded, and entropy evolves smoothly. The comparison supports the role of saturated-group filtering as a training-stability device for continuous dimension-level GRPO.

\subsection{Analysis of Generated Rubrics}
\label{sec:appendix_rubric_analysis}

We conduct a rubric-only analysis on MMRB2 to examine whether the generated rubrics remain stable under stochastic decoding while adapting to different instructions. For each instance, we sample five rubrics with temperature $0.7$ and top-$p=0.9$. For each rubric pair, we first match evaluation dimensions by semantic similarity and then measure dimension similarity and weight difference. Higher dimension similarity indicates more similar evaluation criteria, whereas lower weight difference indicates more consistent importance assigned to the matched criteria.

\begin{table}[t]
\centering
\small
\setlength{\tabcolsep}{2.5pt}
\begin{tabular*}{\columnwidth}{@{\extracolsep{\fill}}lcccc@{}}
\toprule
Task & Same inst. & Same type & Diff. type & Weight diff. \\
\midrule
Gen  & 0.763 & 0.597 & 0.563 & 0.102 \\
Edit & 0.800 & 0.651 & 0.612 & 0.110 \\
\bottomrule
\end{tabular*}
\caption{Stability and adaptation of generated rubrics on MMRB2. ``Same inst.'' denotes the same instruction and ``Diff. type'' denotes different instruction types. The three similarity columns report dimension similarity.}
\label{tab:rubric_stability}
\end{table}

\paragraph{Stability.}
As shown in Table~\ref{tab:rubric_stability}, repeated rubrics for the same instruction have dimension similarities of $0.763$ for generation and $0.800$ for editing, while their weight differences remain small at $0.102$ and $0.110$, respectively. Thus, stochastic decoding does not produce arbitrary evaluation criteria: both the selected dimensions and their relative importance remain largely consistent.

\paragraph{Diversity and adaptation.}
For each instruction, we compare its rubrics with those produced for other prompts of the same instruction type and for prompts of different types. Similarity follows the expected order for both tasks: same instruction $>$ same type $>$ different type. Rubrics therefore preserve shared evaluation patterns among related instructions while varying across instruction types. The higher same-type than different-type similarity for generation ($0.597$ vs. $0.563$) and editing ($0.651$ vs. $0.612$) further indicates instruction-type adaptation. Qualitative inspection likewise shows that the selected evaluation dimensions change according to task requirements.

\subsection{Sensitivity to Teacher Choice}
\label{sec:appendix_teacher_sensitivity}

The teacher model in our pipeline is not an independent preference annotator. Each image-pair preference label is inherited from the original human-annotated data, and the teacher only converts that fixed preference into a rubric trajectory. Teacher bias can therefore affect dimension selection and granularity, but it does not determine the underlying preference label.

To quantify the remaining sensitivity, we conduct a controlled cross-teacher analysis on a fixed MMRB2 subset containing 200 generation and 200 editing comparisons. For every comparison, we keep the input and human preference label fixed and regenerate rubrics with three teacher families: Gemini 3.1 Pro, Claude Sonnet 5, and GPT-5.6 Luna. For each teacher pair, we semantically match the rubric dimensions and report whether at least one dimension is shared, whether at least half of the smaller dimension set is covered, and the average overlap ratio.

\begin{center}
\centering
\small
\setlength{\tabcolsep}{3pt}
\begin{tabular}{lccc}
\toprule
Teacher pair & \makecell{At least one\\shared} & \makecell{At least half\\covered} & \makecell{Average\\overlap} \\
\midrule
Claude vs. Gemini & 93.0\% & 58.5\% & 0.540 \\
Claude vs. GPT    & 98.0\% & 80.0\% & 0.608 \\
Gemini vs. GPT    & 83.5\% & 47.5\% & 0.453 \\
\midrule
Average           & 91.5\% & 62.0\% & 0.534 \\
\bottomrule
\end{tabular}
\captionof{table}{Cross-teacher rubric overlap on a fixed MMRB2 subset. Inputs and human preference labels are held constant across teacher families.}
\label{tab:teacher_sensitivity}
\end{center}

Across teacher pairs, $91.5\%$ of comparisons share at least one semantic dimension, $62.0\%$ cover at least half of the smaller rubric, and the average overlap is $0.534$. Because the generated rubrics generally contain only three to four dimensions, these results indicate substantial agreement across teacher families. Complete agreement is neither observed nor expected: image preference evaluation is inherently subjective, and different valid criteria can support the same human preference. Overall, the teachers recover a shared evaluation core while retaining reasonable differences in criterion selection and granularity.

\section{Prompt Templates}
\label{sec:appendix_prompts}

This appendix provides the inference prompt templates used by RubricRM. Sections~\ref{sec:appendix_prompt_t2i} and~\ref{sec:appendix_prompt_edit} specify the required output structure, including task analysis, dynamic dimension selection, dimension-level scoring, weighted aggregation, and final preference selection.
\onecolumn
\tcbset{
  rubricraw/.style={
    enhanced,
    breakable,
    listing only,
    listing engine=listings,
    colback=black!4,
    colframe=black!70,
    colbacktitle=black!75,
    coltitle=white,
    fonttitle=\bfseries,
    boxrule=0.8pt,
    arc=3pt,
    left=2mm,
    right=2mm,
    top=1mm,
    bottom=1mm,
    listing options={
      basicstyle=\rubriccodefont\fontsize{6.4pt}{7.25pt}\selectfont,
      inputencoding=utf8,
      extendedchars=true,
      breaklines=true,
      breakatwhitespace=false,
      breakautoindent=false,
      breakindent=0pt,
      columns=fullflexible,
      keepspaces=true,
      showstringspaces=false,
      literate=*{→}{{$\rightarrow$}}1
    }
  }
}

\newtcblisting{rubricrawbox}[1]{rubricraw,title={#1},title after break={#1 (continued)}}

\subsection{Text-to-Image Preference Evaluation}
\label{sec:appendix_prompt_t2i}

\begin{rubricrawbox}{Text-to-Image Generation}
# Role
You are an expert evaluator for text-to-image (T2I) generation, specializing in multi-dimensional image assessment. You are skilled at identifying the most relevant evaluation dimensions, assigning appropriate weights, and delivering precise, logically grounded comparisons.

# Workflow
1. Intent Mining: Analyze the prompt to identify its core objective, including required content, key attributes, stylistic constraints, compositional requirements, and any other critical conditions, while also taking into account general image evaluation principles.
2. Dimension Selection & Weighting: Select 3-5 evaluation dimensions that are most relevant to the task. All dimensions must be atomic: each dimension should assess exactly one distinct aspect and must not combine multiple criteria into a single dimension. Assign weights dynamically based on their importance. The total weight must sum to 100%.
3. Dimension-based Scoring: Evaluate each image on every selected dimension using the following 0-4 rubric. Scores must be assigned relative to the specific dimension being evaluated.
   0: Failed: Does not satisfy the dimension; severe errors or breakdowns are present.
   1: Poor: Satisfies the dimension only weakly; major deviations, omissions, or artifacts are present.
   2: Fair: Partially satisfies the dimension; the intended quality is present, but notable issues remain.
   3: Good: Satisfies the dimension well; only minor flaws are present.
   4: Excellent: Fully satisfies the dimension; highly consistent and essentially free of noticeable flaws.

# Output Format
## [Thinking Process]
- Task Analysis: Systematically analyze the prompt.
- Selected Dimensions & Weights: List the chosen dimensions and explain why each one is important, including the rationale for its assigned weight.

## [Detailed Evaluation]
- [Dimension Name] ([Weight]%)
  - Image A: [brief analysis] → Score: X/4
  - Image B: [brief analysis] → Score: X/4

## [Final Conclusion]
- Weighted Total Score: For each image, show the weighted score calculation and provide the final result rounded to 2 decimal places.
- Summary: Briefly summarize the key reasons behind the evaluation.
- Preference: You **must** output exactly one answer -- \boxed{{A}} or \boxed{{B}}. A definitive choice is required even when scores are tied.

# Prompt
{prompt}

# Images
- Image A: <image>
- Image B: <image>
\end{rubricrawbox}

\clearpage

\subsection{Image-Editing Preference Evaluation}
\label{sec:appendix_prompt_edit}

\begin{rubricrawbox}{Image Editing}
# Role
You are an expert evaluator for image editing, specializing in multi-dimensional assessment of edited images. You are skilled at identifying the most relevant evaluation dimensions, assigning appropriate weights, and delivering precise, logically grounded comparisons.

# Workflow
1. Intent Mining: Analyze the editing instruction and original image to identify its core objective, including required edits, key attributes, stylistic constraints, content preservation requirements, and any other critical conditions, while also taking into account general image evaluation principles.
2. Dimension Selection & Weighting: Select 3-5 evaluation dimensions that are most relevant to the task. All dimensions must be atomic: each dimension should assess exactly one distinct aspect and must not combine multiple criteria into a single dimension. Assign weights dynamically based on their importance. The total weight must sum to 100%.
3. Dimension-based Scoring: Evaluate each image on every selected dimension using the following 0-4 rubric. Scores must be assigned relative to the specific dimension being evaluated.
   0: Failed: Does not satisfy the dimension; severe errors or breakdowns are present.
   1: Poor: Satisfies the dimension only weakly; major deviations, omissions, or artifacts are present.
   2: Fair: Partially satisfies the dimension; the intended quality is present, but notable issues remain.
   3: Good: Satisfies the dimension well; only minor flaws are present.
   4: Excellent: Fully satisfies the dimension; highly consistent and essentially free of noticeable flaws.

# Output Format
## [Thinking Process]
- Task Analysis: Systematically analyze the editing instruction.
- Selected Dimensions & Weights: List the chosen dimensions and explain why each one is important, including the rationale for its assigned weight.

## [Detailed Evaluation]
- [Dimension Name] ([Weight]%)
  - Image A: [brief analysis] → Score: X/4
  - Image B: [brief analysis] → Score: X/4

## [Final Conclusion]
- Weighted Total Score: For each image, show the weighted score calculation and provide the final result rounded to 2 decimal places.
- Summary: Briefly summarize the key reasons behind the evaluation.
- Preference: You **must** output exactly one answer -- \boxed{{A}} or \boxed{{B}}. A definitive choice is required even when scores are tied.

# Editing Instruction
{prompt}

# Images
- Image A: <image>
- Image B: <image>
\end{rubricrawbox}

\newpage
\section{Running Cases}
\label{sec:appendix_running_cases}
\subsection{Text-to-Image Generation}
\label{sec:appendix_running_t2i}
\begin{rubricrawbox}{Input Prompt}
A cranberry bog flooded for harvest with visible red berries, featuring geese flying in V-shaped flocks during autumn.
\end{rubricrawbox}

\begin{center}
\begin{minipage}[t]{0.31\textwidth}
\centering
\includegraphics[width=\linewidth]{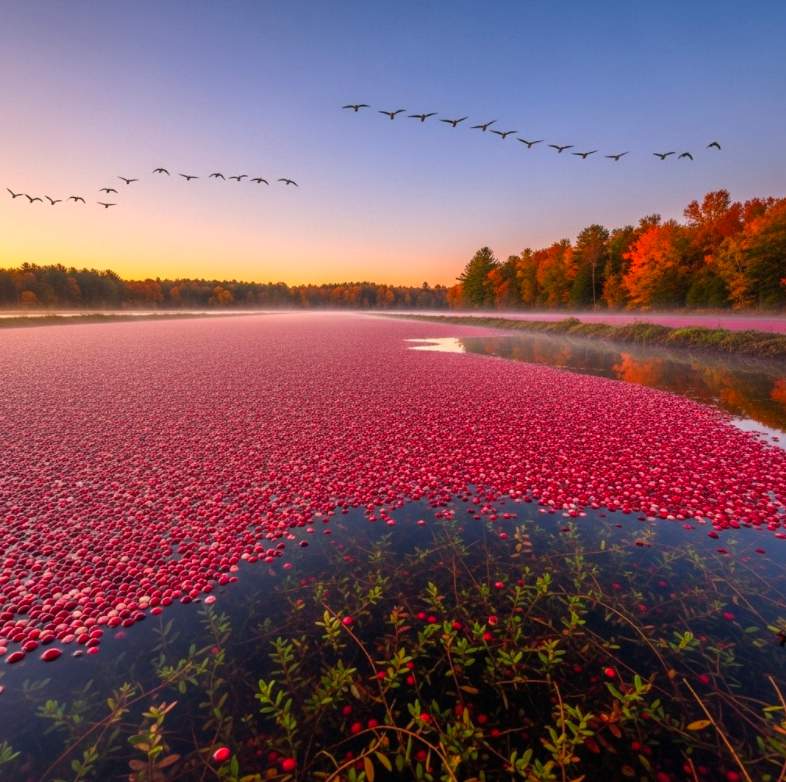}
\vspace{2pt}
\textbf{Image A}
\end{minipage}
\hspace{0.06\textwidth}
\begin{minipage}[t]{0.31\textwidth}
\centering
\includegraphics[width=\linewidth]{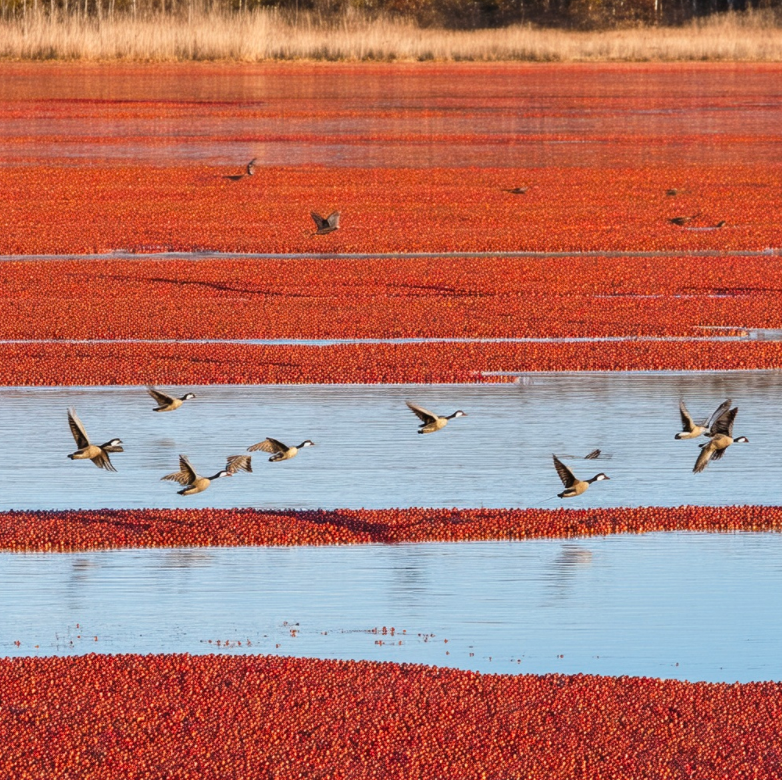}
\vspace{2pt}
\textbf{Image B}
\end{minipage}
\end{center}

\begin{rubricrawbox}{Model Output}
## [Thinking Process]

- **Task Analysis**: The prompt requires a specific agricultural scene: a flooded cranberry bog during harvest. Key elements include visible red berries, geese flying in a V-shape, and an autumn setting. The goal is to capture the unique "harvest" look of cranberries (which float when flooded) while maintaining naturalistic details and seasonal atmosphere.
- **Selected Dimensions & Weights**:
    1. **Prompt Adherence (Content)** (30%): Evaluates the presence of all required elements: flooded bog, red berries, and geese in a V-shape.
    2. **Visual Realism & Detail** (30%): Assesses the quality of the rendering, specifically the texture of the berries and the anatomical correctness of the geese.
    3. **Composition & Aesthetics** (20%): Evaluates the layout, lighting, and how well the "autumn" atmosphere is conveyed.
    4. **Logical Consistency** (20%): Checks for AI artifacts, such as impossible geometry or illogical placement of objects (e.g., berries in the water vs. on the ground).

## [Detailed Evaluation]

- **Prompt Adherence (Content)** (30%)
  - Image A: Includes a flooded bog, visible red berries, and geese in a V-shape. The autumn setting is clear. → Score: 4/4
  - Image B: Includes a flooded bog, red berries, and geese. However, the geese are not in a clear V-shape, and the "flooded" aspect is rendered as strange horizontal stripes. → Score: 2/4

- **Visual Realism & Detail** (30%)
  - Image A: The berries have a realistic texture and scale. The geese are silhouetted but clearly defined. The lighting is natural. → Score: 4/4
  - Image B: The berries look like a repetitive, low-resolution texture. The geese are poorly rendered, with distorted wings and unnatural poses. → Score: 1/4

- **Composition & Aesthetics** (20%)
  - Image A: Excellent use of perspective and lighting. The sunset/autumn colors in the trees provide a beautiful contrast to the red berries. → Score: 4/4
  - Image B: The composition is very flat and repetitive. The horizontal banding of the bog is visually jarring and lacks depth. → Score: 1/4

- **Logical Consistency** (20%)
  - Image A: The scene is logically consistent. The berries are floating on the surface as they would during a harvest. → Score: 4/4
  - Image B: The image contains significant logical errors. The horizontal stripes of berries and water are illogical for a flooded bog. Some geese appear to be floating or walking on the water surface in a physically impossible way. → Score: 0/4

## [Final Conclusion]

- **Weighted Total Score**:
  - Image A: (4 * 0.30) + (4 * 0.30) + (4 * 0.20) + (4 * 0.20) = **4.00**
  - Image B: (2 * 0.30) + (1 * 0.30) + (1 * 0.20) + (0 * 0.20) = **1.10**

- **Summary**: Image A is a closer realization of the prompt. It captures the specific agricultural look of a cranberry harvest with realistic details, autumn lighting, and logical composition. Image B is weaker in realism and logic, featuring repetitive textures, distorted birds, and an illogical "striped" interpretation of a flooded bog.

- **Preference**: \boxed{A}
\end{rubricrawbox}

\clearpage
\subsection{Image Editing}
\label{sec:appendix_running_edit}
\begin{rubricrawbox}{Editing Instruction}
Create a colorful educational poster for a kindergarten classroom using the toy airplane image as the centerpiece. Add the title: "Let's Learn About Airplanes!" at the top. On the left side, include a fun fact: "Airplanes travel high in the sky and can go really fast!" On the right side, add a safety message: "Always fasten your seatbelt when flying!" At the bottom, write: "Can you name the colors of this plane?" in bold letters.
\end{rubricrawbox}

\begin{center}
\begin{minipage}[t]{0.31\textwidth}
\centering
\includegraphics[width=\linewidth]{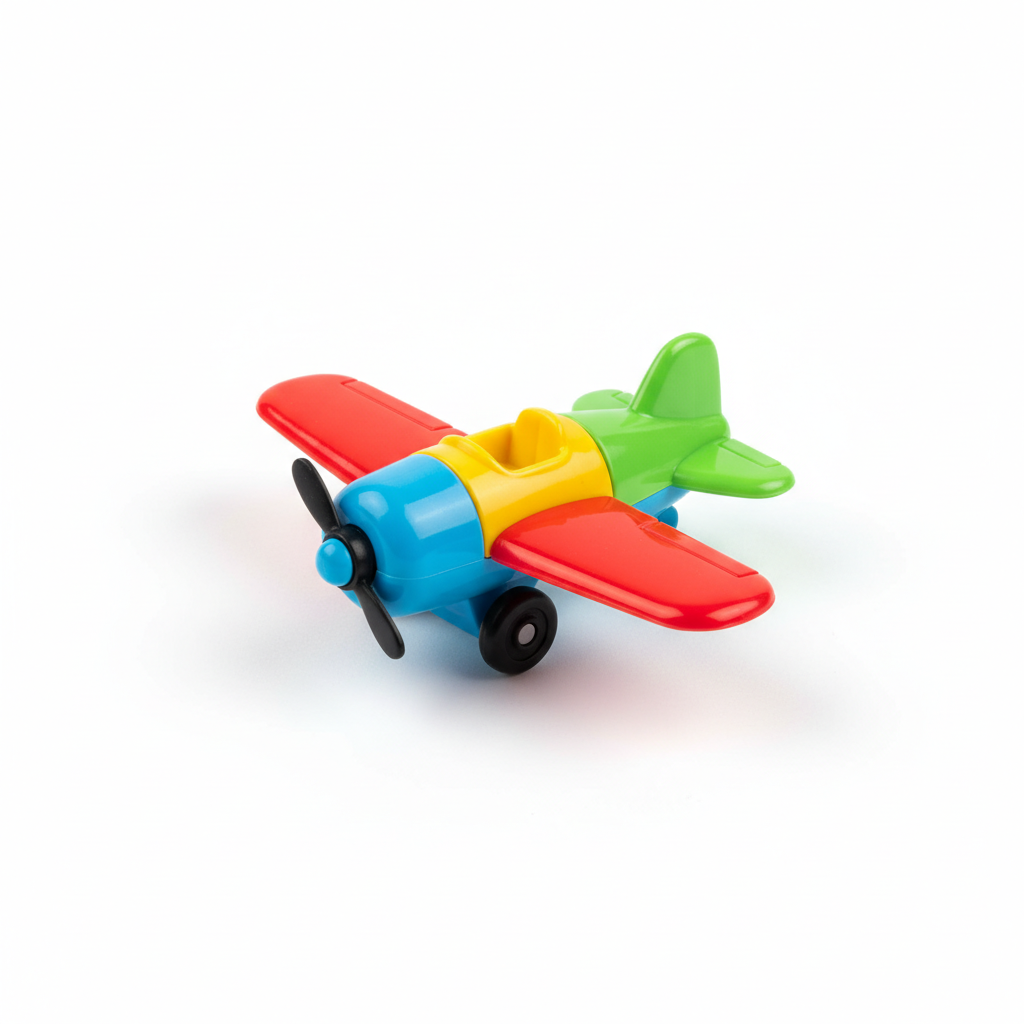}
\vspace{2pt}
\textbf{Original Image}
\end{minipage}
\hfill
\begin{minipage}[t]{0.31\textwidth}
\centering
\includegraphics[width=\linewidth]{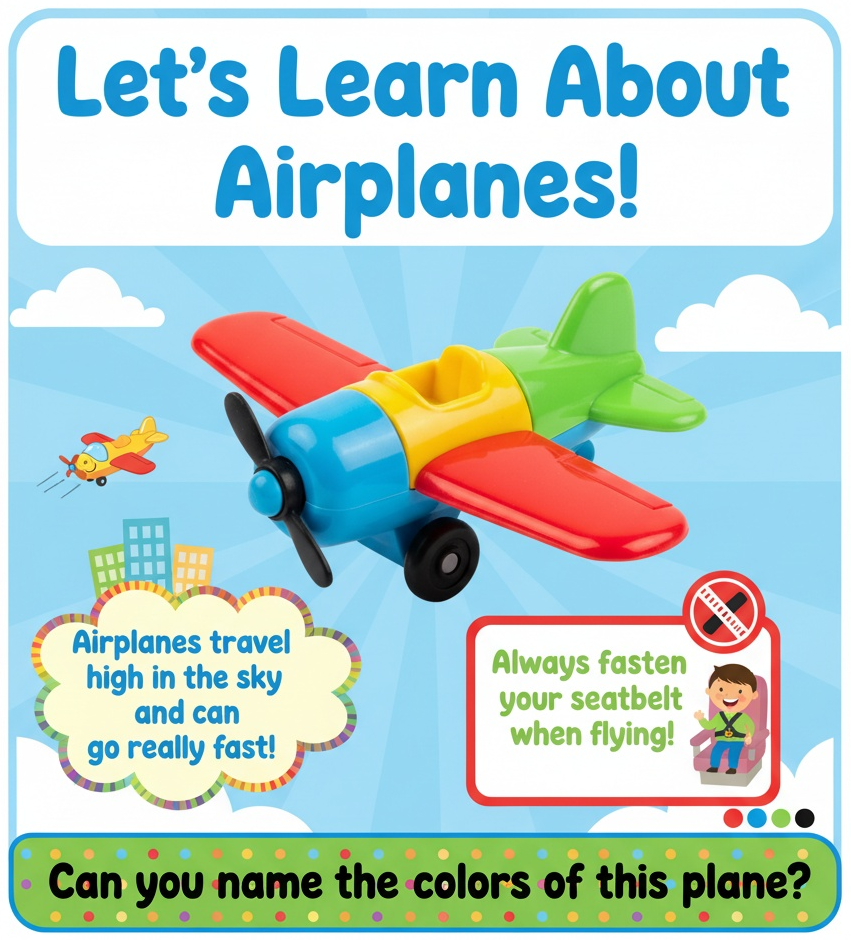}
\vspace{2pt}
\textbf{Image A}
\end{minipage}
\hfill
\begin{minipage}[t]{0.31\textwidth}
\centering
\includegraphics[width=\linewidth]{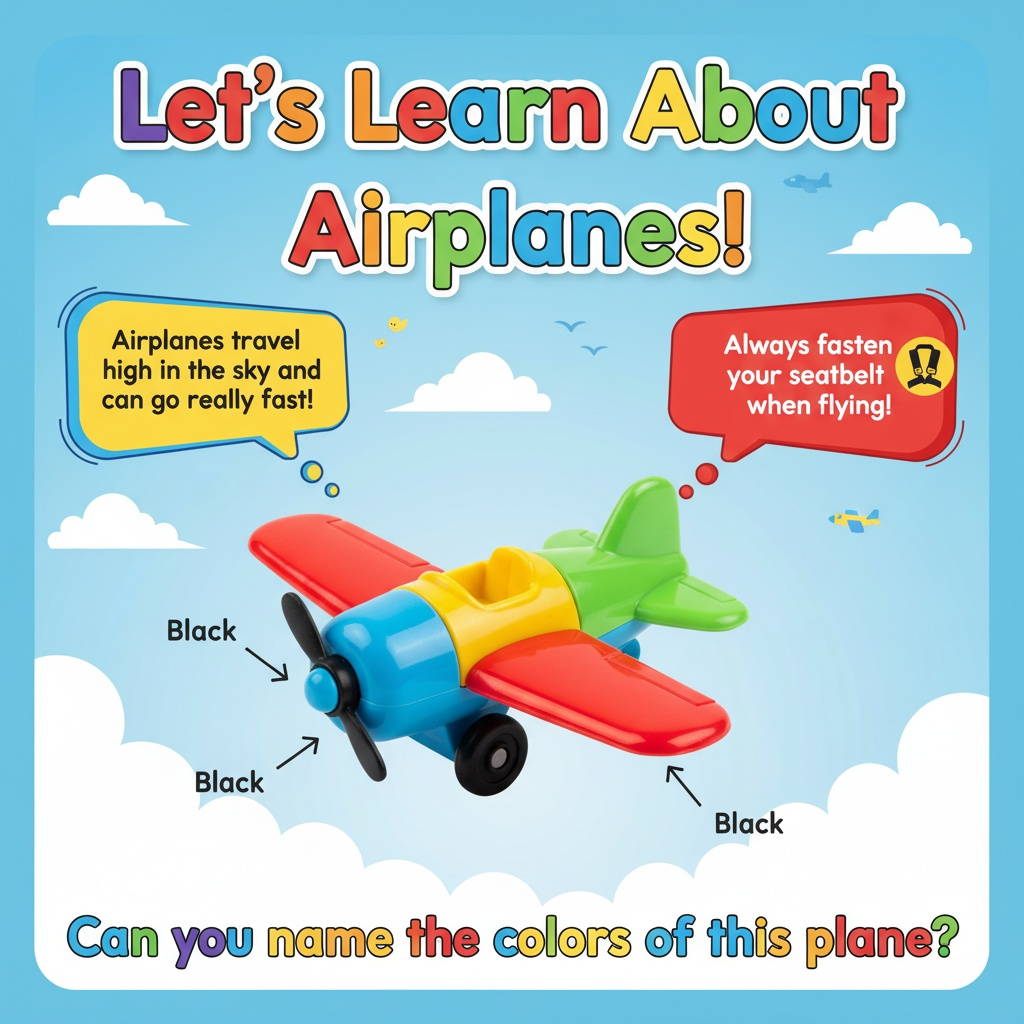}
\vspace{2pt}
\textbf{Image B}
\end{minipage}
\end{center}

\begin{rubricrawbox}{Model Output}
## [Thinking Process]
- **Task Analysis**: The user wants to transform a simple image of a toy airplane into a colorful educational poster for a kindergarten classroom. The instruction includes several strict requirements:
  1.  **Style**: Colorful, educational poster for a kindergarten classroom.
  2.  **Centerpiece**: The original toy airplane image.
  3.  **Text & Placement**:
      *   Top: "Let's Learn About Airplanes!"
      *   Left side: "Airplanes travel high in the sky and can go really fast!"
      *   Right side: "Always fasten your seatbelt when flying!"
      *   Bottom: "Can you name the colors of this plane?" in bold letters.
- **Selected Dimensions & Weights**:
  1.  **Text Accuracy & Placement (40%)**: This is the most complex and specific part of the prompt. It evaluates whether all four text elements are present, spelled correctly, and placed in the exact requested locations.
  2.  **Thematic Style & Aesthetics (30%)**: Evaluates whether the overall design successfully captures the "colorful educational poster for a kindergarten classroom" vibe.
  3.  **Content Preservation (30%)**: Evaluates how well the original toy airplane is preserved as the centerpiece, without introducing unwanted artifacts or altering its core identity.

## [Detailed Evaluation]
- **Text Accuracy & Placement (40%)**
  - Image A: All four text elements are present, spelled correctly, and placed exactly where requested (top, left, right, bottom). The bottom text is also bold, as requested. → Score: 4/4
  - Image B: All four text elements are present and spelled correctly. The text is placed in the correct locations. The bottom text is bold. → Score: 4/4

- **Thematic Style & Aesthetics (30%)**
  - Image A: The image features a bright, colorful, and playful design with clouds, a sunburst background, and kid-friendly graphics. It perfectly captures the requested kindergarten classroom poster aesthetic. → Score: 4/4
  - Image B: The image is colorful and features a sky background, but the overall aesthetic feels a bit more generic and less cohesive as a kindergarten poster compared to Image A. → Score: 3/4

- **Content Preservation (30%)**
  - Image A: The original toy airplane is perfectly preserved in the center. The image also includes some nice, non-distracting background elements (like a small cartoon plane and city skyline) that fit the theme. → Score: 4/4
  - Image B: The original toy airplane is preserved, but the model added strange, incorrect text labels ("Black") pointing to the propeller and wings, which detracts from the original image's integrity. → Score: 2/4

## [Final Conclusion]
- **Weighted Total Score**:
  - Image A: (4 * 0.40) + (4 * 0.30) + (4 * 0.30) = 1.60 + 1.20 + 1.20 = 4.00
  - Image B: (4 * 0.40) + (3 * 0.30) + (2 * 0.30) = 1.60 + 0.90 + 0.60 = 3.10
- **Summary**: Both images successfully followed the strict text and placement instructions. However, Image A is vastly superior in its overall aesthetic, creating a much more engaging and appropriate kindergarten poster. Furthermore, Image A perfectly preserves the original toy airplane, whereas Image B introduces unwanted and incorrect text labels pointing to the plane.
- **Preference**: \boxed{A}
\end{rubricrawbox}

\end{document}